\documentclass[10pt,twocolumn,letterpaper]{article}

\usepackage{iccv}
\usepackage{times}
\usepackage{epsfig}
\usepackage{graphicx}
\usepackage{amsmath}
\usepackage{amssymb}
\usepackage{lipsum}
\usepackage{multirow}
\usepackage{subfigure}
\usepackage{color}
\usepackage{overpic}
\usepackage{caption}

\usepackage[pagebackref=true,breaklinks=true,letterpaper=true,colorlinks,bookmarks=false]{hyperref}

\def\up#1{$_{(\textcolor[rgb]{0,0.75,0.25}{+{#1}})}$}

\newlength\savewidth\newcommand\shline{\noalign{\global\savewidth\arrayrulewidth
  \global\arrayrulewidth 1pt}\hline\noalign{\global\arrayrulewidth\savewidth}}

\iccvfinalcopy 

\ificcvfinal\fi

\begin{document}

\title{Change is Everywhere: Single-Temporal Supervised Object Change Detection \\ in Remote Sensing Imagery}

\author{Zhuo Zheng\qquad Ailong Ma\qquad Liangpei Zhang\qquad Yanfei Zhong\thanks{corresponding author.} \\
Wuhan University, Wuhan, China\\
{\tt\small \{zhengzhuo, maailong007, zlp62, zhongyanfei\}@whu.edu.cn}
}

\maketitle
\ificcvfinal\thispagestyle{empty}\fi

\begin{abstract}
For high spatial resolution (HSR) remote sensing images, bitemporal supervised learning always dominates change detection using many pairwise labeled bitemporal images.
However, it is very expensive and time-consuming to pairwise label large-scale bitemporal HSR remote sensing images.
In this paper, we propose single-temporal supervised learning (STAR) for change detection from a new perspective of exploiting object changes in unpaired images as supervisory signals.
STAR enables us to train a high-accuracy change detector only using \textbf{unpaired} labeled images and generalize to real-world bitemporal images.
To evaluate the effectiveness of STAR, we design a simple yet effective change detector called ChangeStar, which can reuse any deep semantic segmentation architecture by the ChangeMixin module.
The comprehensive experimental results show that ChangeStar outperforms the baseline with a large margin under single-temporal supervision and achieves superior performance under bitemporal supervision.
Code is available at \url{https://github.com/Z-Zheng/ChangeStar}.

\end{abstract}

\vspace{-0.1in}
\section{Introduction}
\label{sec:intro}
Object change detection using multi-temporal high spatial resolution (HSR) remote sensing imagery is a meaningful but challenging fundamental task in remote sensing and earth vision, which can provide more accurate object change information of land surface for urban expansion, urban planning, environmental monitoring, and disaster assessment \cite{hussain2013change, zhang2017separate, daudt2018urban, mahdavi2019polsar, gupta2019creating}.
This task takes bitemporal images as input and outputs pixel-wise object change.

\begin{figure}
\centering
\subfigure[Conventional Bitemporal Supervised Learning]{
    \begin{minipage}[b]{\linewidth}
        \includegraphics[width=\linewidth]{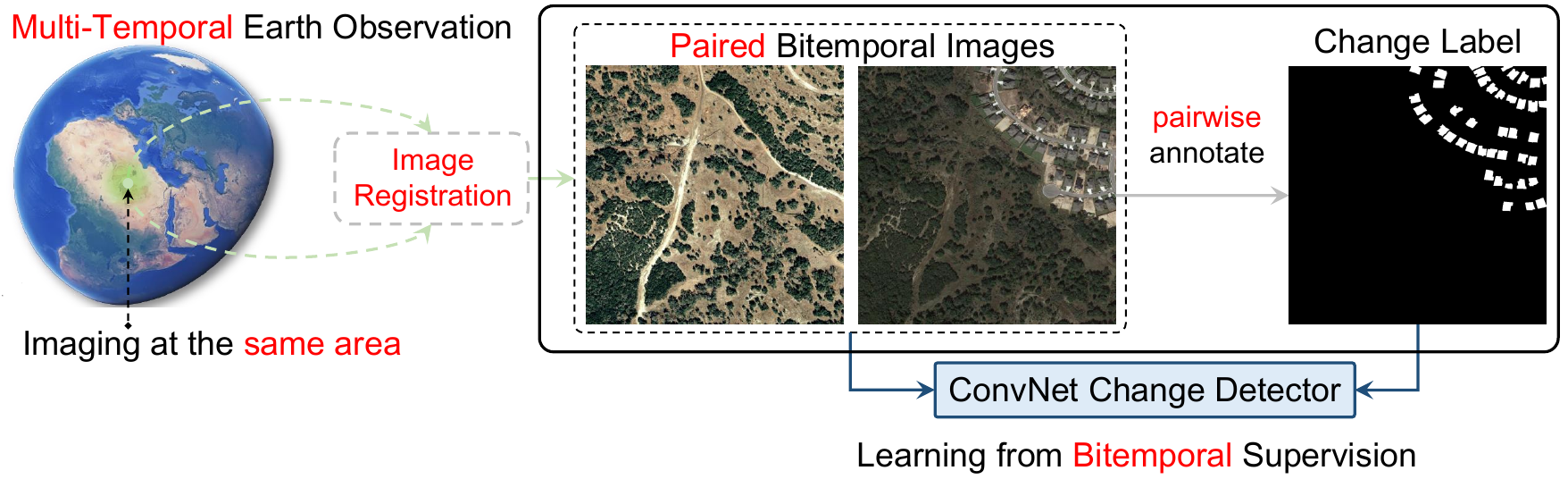}
    \end{minipage}
}

\subfigure[STAR: Single-Temporal supervised leARning]{
    \begin{minipage}[b]{\linewidth}
        \includegraphics[width=\linewidth]{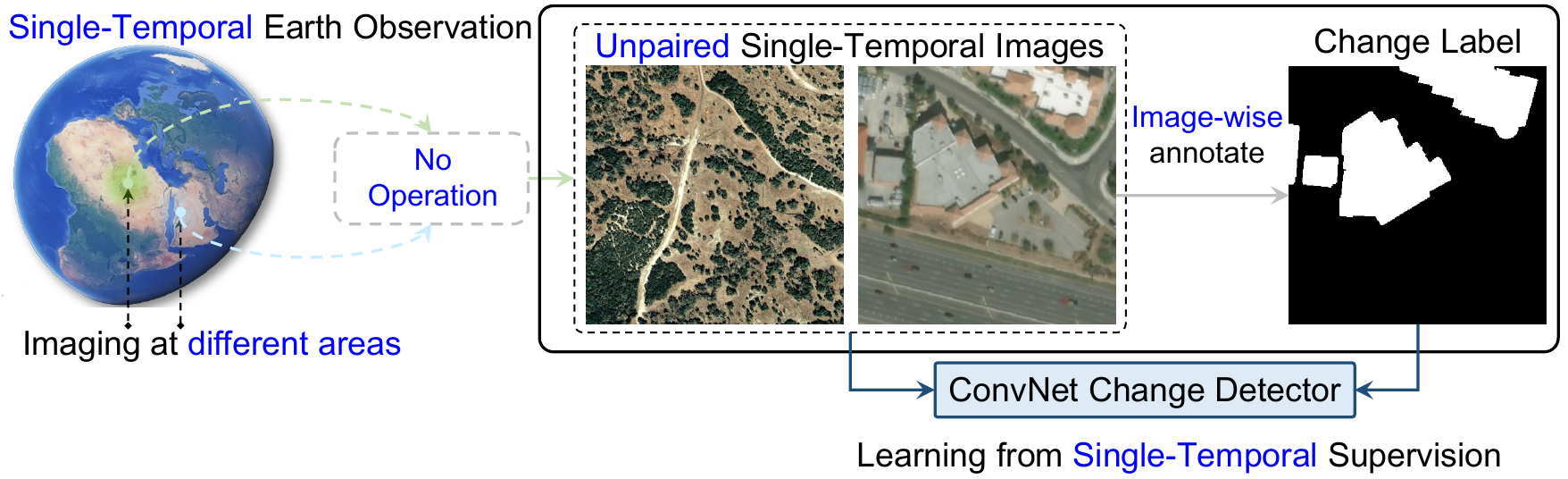}
    \end{minipage}
}
\caption{Comparison of conventional bitemporal supervised learning and the proposed single-temporal supervised learning for object change detection.
By exploiting object changes in arbitrary image pairs as the supervisory signals, STAR makes it possible to learn a change detector from unpaired single-temporal images.
}
\label{fig:comp}
\end{figure}

The dominating change detection methods are based on deep convolutional neural networks (ConvNet) toward high-accuracy and reliable geospatial object change detection in complex application scenarios.
Learning a ConvNet change detector needs a large number of pairwise labeled bitemporal images with bitemporal supervision, as shown in Fig.~\ref{fig:comp} (a).
However, pairwise labeling large-scale and high-quality bitemporal HSR remote sensing images is very expensive and time-consuming because of the extensive coverage of remote sensing images. 
This significantly limits the real-world applications of the change detection technique.

We observed that the importance of pairwise labeled bitemporal images lies in that the change detector needs paired semantic information to define positive and negative samples for object change detection.
These positive and negative samples are usually determined by whether the pixels at two different times have different semantics in the same geographical area. 
The semantics of bitemporal pixels controls the label assignment, while the positional consistency condition\footnote{The bitemporal pixels should be at the same geographical position.} is only used to guarantee independent and identically distributed (i.i.d.) training and inference.
It is conceivable that change is everywhere, especially between unpaired images, if we relax the positional consistency condition to define positive and negative samples.

In this paper, we propose a single-temporal supervised object change detection approach to bypass the problem of collecting paired labeled images by exploiting object change between unpaired images as supervisory signals, as shown in Fig.~\ref{fig:comp} (b).
This approach enables us to train a high-accuracy change detector using unpaired labeled images and generalize to real-world bitemporal images at the inference stage.
Because it only needs single-temporal semantic segmentation labels to construct object changes as change detection labels, we refer to our approach as \textit{Single-Temporal supervised leARning} (STAR).

Conditioned by the same geographical area, bitemporal supervised learning can avoid many out-of-distribution positive samples, whereas this is both an opportunity and a challenge for the STAR.
These out-of-distribution samples make the change detector driven by STAR more potential to possess better generalization.
Meanwhile, they also cause the overfitting problem to make the model learn biased representation.
To alleviate this problem, we explore an inductive bias: temporal symmetry and leverage it to constraint the representation learning for the change detector.

To demonstrate the effectiveness of STAR algorithm, we design a simple yet unified change detector called \textit{ChangeStar}, which follows the modular design and is made up of an arbitrary deep semantic segmentation model and the ChangeMixin module driven by STAR.
The ChangeMixin module is designed to enable an arbitrary deep semantic segmentation model to detect object change.
This allows ChangeStar to reuse excellent semantic segmentation architectures to assist in change detection without extra specific architecture design, which bridges the gap between semantic segmentation and change detection.

The main contributions of this paper are summarized as follows:
\begin{itemize}
    \vspace{-0.1in}
    \item To fundamentally alleviate the problem of collecting paired labeled images, we proposed single-temporal supervised learning (STAR) to enable object change detectors to learn from unpaired labeled images.
    \vspace{-0.1in}
    \item To further stabilize the learning, we explore and leverage an inductive bias, temporal symmetry, to alleviate the overfitting problem caused by the absence of positional consistency condition in unpaired images.
    \vspace{-0.1in}
    \item To reuse the modern semantic segmentation architectures, we proposed a simple yet effective multi-task architecture, called ChangeStar, for joint semantic segmentation and change detection.
    The core component of ChangeStar is the ChangeMixin, which enables off-the-shelf segmentation model to detect object change.
\end{itemize}

\section{Related Work}
\label{sec:related_work}

\paragraph{Object Change Detection.}
Different from general remote sensing change detection \cite{singh1989review}, object change detection is an object-centric change detection, which is aimed at answering the question of whether the object of interest has been changed. 
By the type of change, object change detection can be divided into two categories: binary object change detection, i.e. building change detection \cite{ji2018fully, chen2020spatial}, and semantic object change detection, i.e. building damage assessment \cite{gupta2019creating}, land cover change detection \cite{tian2020hiucd}.
Binary object change detection is a fundamental problem for object change detection.
Thus, we focus on binary object change detection in this work.

\vspace{-0.2in}
\paragraph{Bitemporal Supervised Learning.}
So far the supervised object change detection methods are based on bitemporal supervised learning, which needs change labels from bitemporal remote sensing images of the same area.
Although there are many change detection benchmark datasets \cite{benedek2009change,bourdis2011constrained,fujita2017damage, lebedev2018change, ji2018fully, daudt2018urban, daudt2019multitask, chen2020spatial, tian2020hiucd}, their scales are still limited for meeting deep learning model.
Because the pairwise annotation is very expansive and time-consuming.
Therefore, a more label-efficient learning algorithm for the change detector is necessary for real-world applications.

\vspace{-0.2in}
\paragraph{Deep ConvNet Change Detector.}
Towards HSR remote sensing geospatial object change detection, the dominant change detectors are based on deep ConvNet \cite{NIPS2012_c399862d}, especially fully convolutional siamese network (FC-Siam) \cite{daudt2018fully}.
FC-Siam adopted a weight-shared encoder to extract temporal-wise deep features and then used a temporal feature difference decoder to detect object change from the perspective of encoder-decoder architecture.
The further improvements mainly focus on three perspectives of the encoder, i.e. using pretrained deep network as the encoder \cite{chen2020spatial, zhang2020deeply}, the decoder, i.e. RNN-based decoders \cite{mou2018learning, chen2019change}, spatial-temporal attention-based decoders \cite{chen2020spatial,zhang2020deeply}, and the training strategy, i.e. deep supervision for multiple outputs \cite{peng2019end,zhang2020deeply}.
It can be found that there are obvious redundant network architecture designs because these network architectures are motivated by the modern semantic segmentation models.
Therefore, it is significantly important for the next generation change detector to reuse modern semantic segmentation architectures.

\vspace{-0.2in}
\paragraph{Object Segmentation.}
An intuitive yet effective single-temporal supervised object change detection method is the post-classification comparison (PCC), which can serve as a strong baseline with the help of the modern object segmentation model.
However, this method only simply treats the change detection task as the semantic segmentation task and ignores the temporal information modeling, thus significantly decreasing the performance.

\begin{figure}[htb]
  \centering
  \subfigure[$t_1$ image]{
      \begin{minipage}[b]{0.3\linewidth}
          \includegraphics[width=\linewidth]{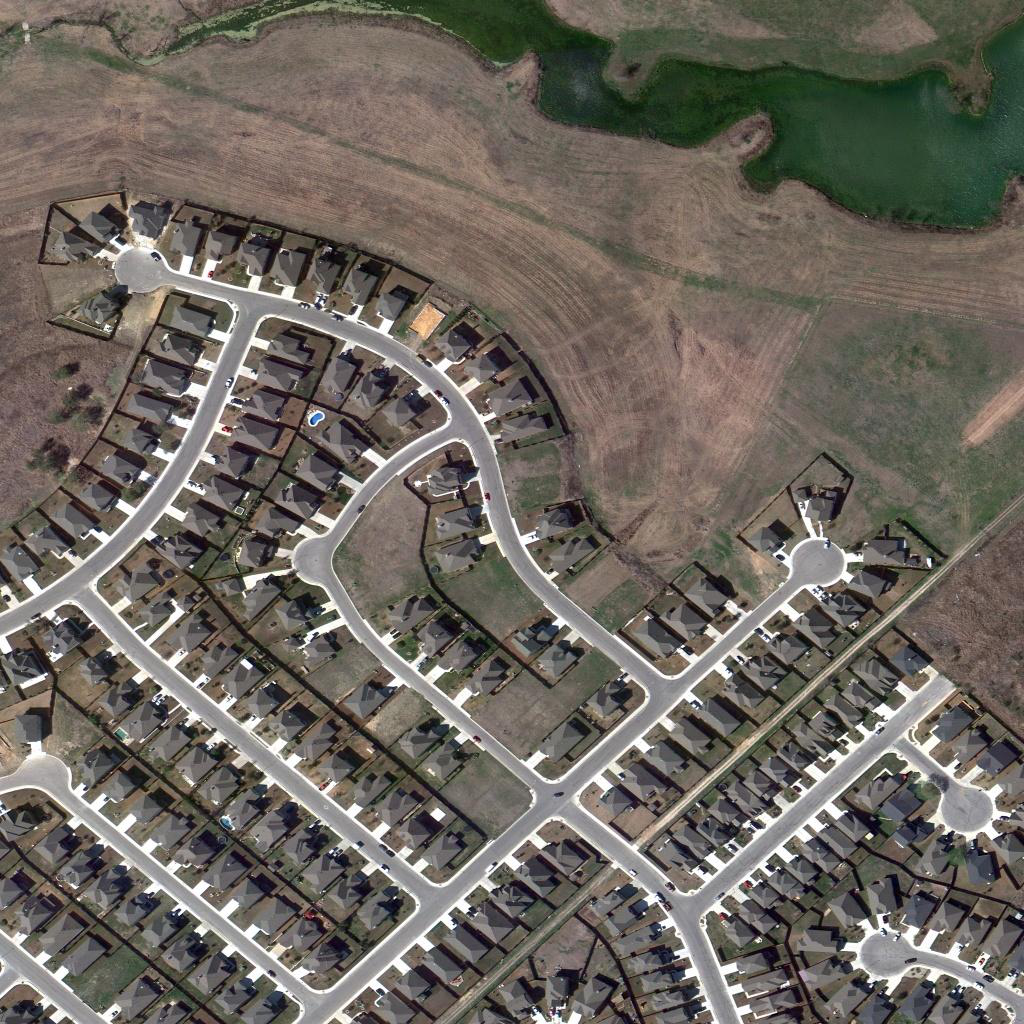}
      \end{minipage}
  }
  \subfigure[$t_2$ image]{
      \begin{minipage}[b]{0.3\linewidth}
          \includegraphics[width=\linewidth]{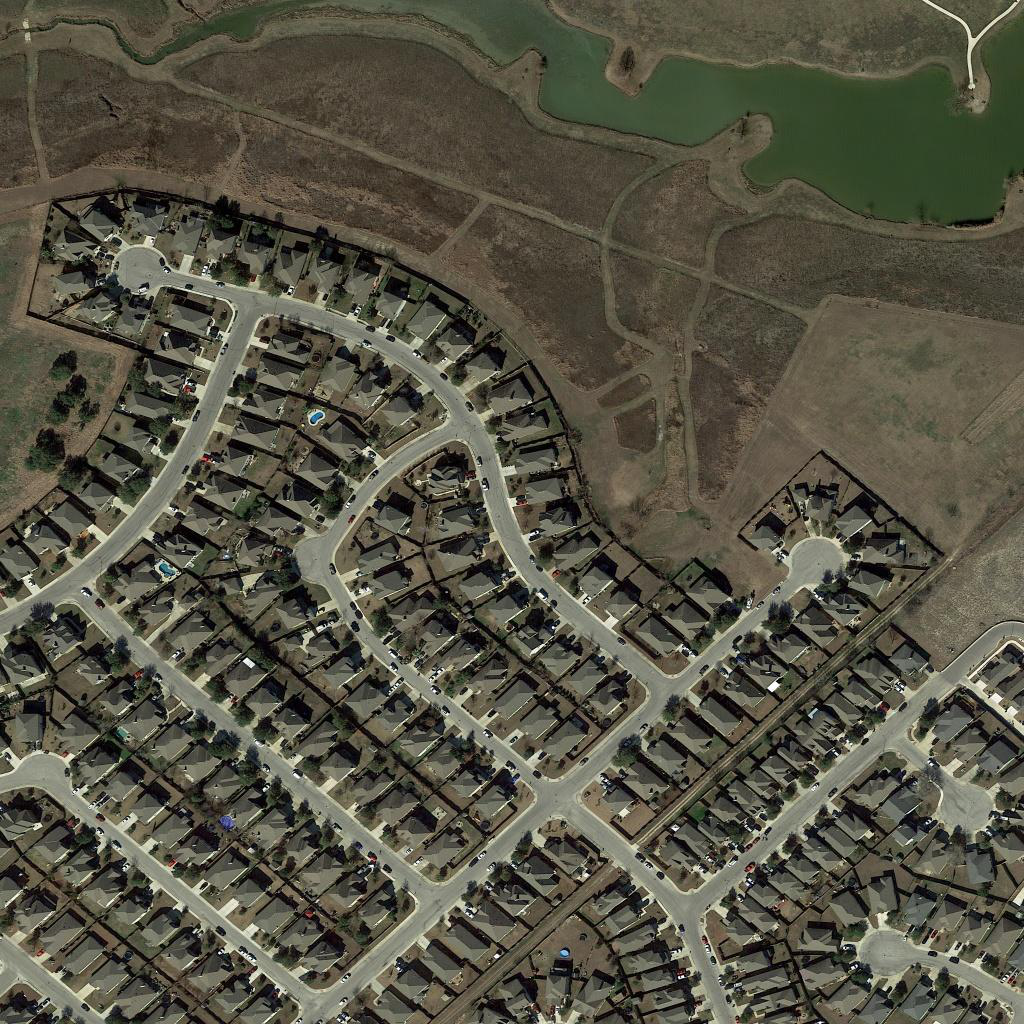}
      \end{minipage}
  }
  \subfigure[$t_1 \rightarrow t_2$ label]{
      \begin{minipage}[b]{0.3\linewidth}
          \includegraphics[width=\linewidth]{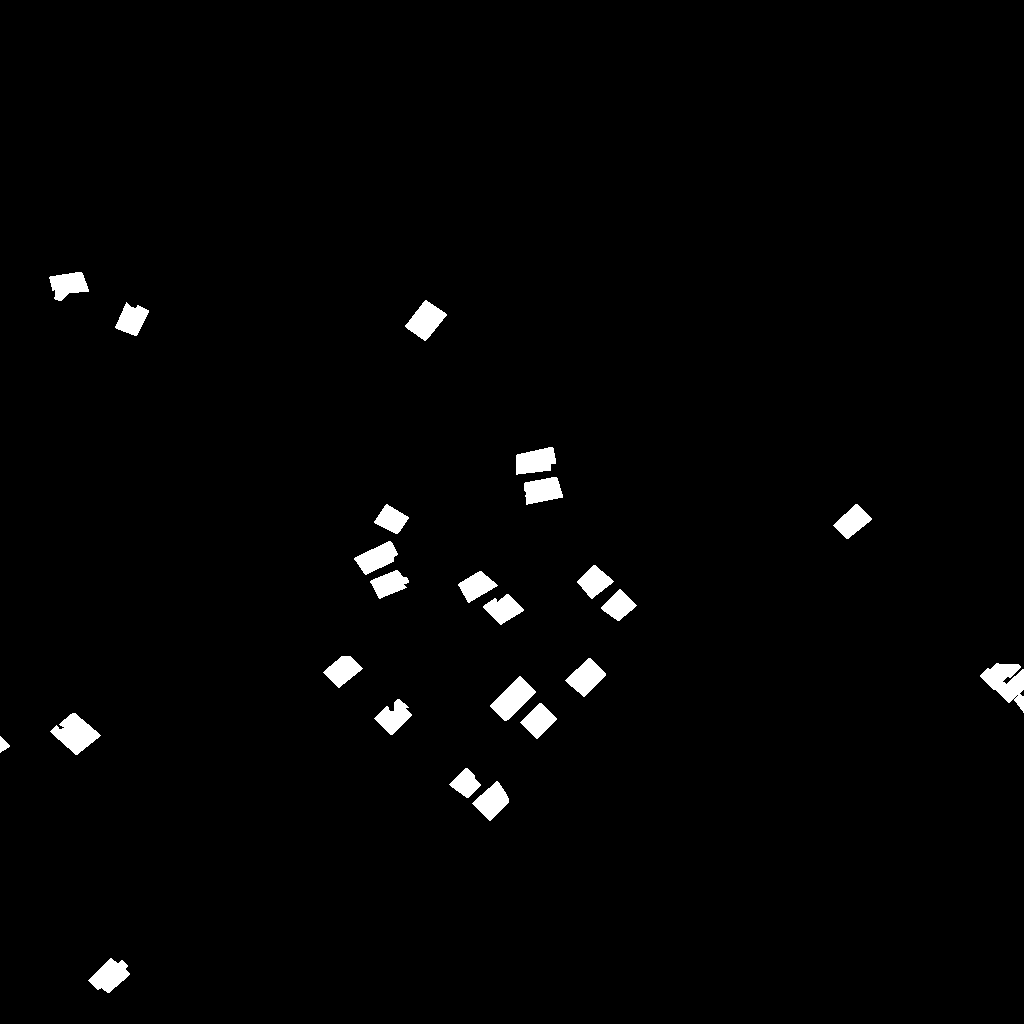}
      \end{minipage}
  }
  \vspace{-0.1in}
  \caption{Training sample of bitemporal supervised object change detection. (a) the image at time $t_1$. (b) the image at time $t_2$. (c) change label representing the change happened the time period from $t_1$ to $t_2$.
  The $t_1$ image must be co-registered with the $t_2$ image for the accurate supervision.
  }
  \label{fig:training_sample}
\end{figure}

\section{Approach}
\label{sec:method}
\subsection{Rethinking Bitemporal Supervised Learning}
Learning an object change detector with bitemporal supervision can be formulated as an optimization problem:
\begin{equation}\label{eqn:org}\vspace{-1mm}
  \mathop{{\rm min}}\limits_{\theta} \, \mathcal{L}(\mathbf{F}_\theta(\mathbf{X}^{t_1}, \mathbf{X}^{t_2}), \mathbf{Y}^{t_1 \rightarrow t_2})
\end{equation}
where $\mathcal{L}$ indicates the objective function that minimizes the cost between the prediction obtained by the object change detector $\mathbf{F}_\theta$ on paired bitemporal images $\mathbf{X}^{t_1}, \mathbf{X}^{t_2} \in \mathbb{R}^{N\times C\times H\times W}$ and change label $\mathbf{Y}^{t_1 \rightarrow t_2} \in \mathbb{R}^{N\times H\times W}$ representing the change happened in the time period from $t_1$ to $t_2$.
For example, Fig.~\ref{fig:training_sample} presents a training sample of bitemporal supervised object change detection.

The core of bitemporal supervised learning is to train a change detector with labeled images at the same spatial position and different times, thus, the training stage is consistent with the inference stage.
From Eq.~\ref{eqn:org}, we can find that change label $\mathbf{Y}^{t_1 \rightarrow t_2}$ is the only source of supervisory signals.
To obtain $\mathbf{Y}^{t_1 \rightarrow t_2}$, paired semantic information is usually needed to define the positive and negative samples. 
However, paired semantic information is only related to the semantics of bitemporal pixels and is unrelated to their spatial positions.
The same spatial position is only used to guarantee the consistency between training and inference.  
If we relax this condition, the original problem in Eq.~\ref{eqn:org} can be simplified as:
\begin{equation}\label{eqn:gen}\vspace{-1mm}
  \mathop{{\rm min}}\limits_{\theta} \, \mathcal{L}(\mathbf{F}_\theta(\mathbf{X}^i, \mathbf{X}^j), {\rm compare}(\mathbf{Y}^i, \mathbf{Y}^j))
\end{equation}
where $\mathbf{X}^i, \mathbf{X}^j$ can be two unpaired images, and supervisory signals are more efficiently collected from semantic comparison between their semantic label $\mathbf{Y}^i, \mathbf{Y}^j$.
The model learned by Eq.~\ref{eqn:gen} is a superset of the model learned by Eq.~\ref{eqn:org}, which is allowed to detect object change in any context, including multi-temporal remote sensing images of the same area.
The original problem can be significantly simplified.

\begin{figure}[htb]
  \centering
  \subfigure[$\mathbf{X}^{t_1}$]{
      \begin{minipage}[b]{0.3\linewidth}
          \includegraphics[width=\linewidth]{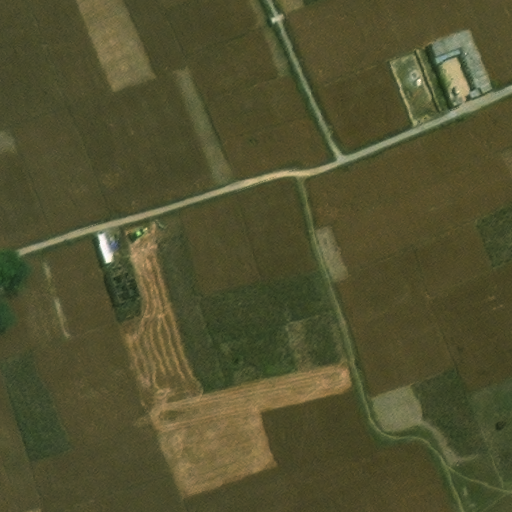}\vspace{4pt}
          \includegraphics[width=\linewidth]{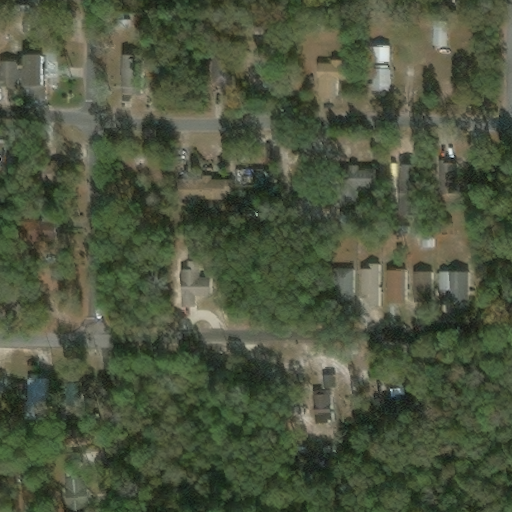}\vspace{4pt}
          \includegraphics[width=\linewidth]{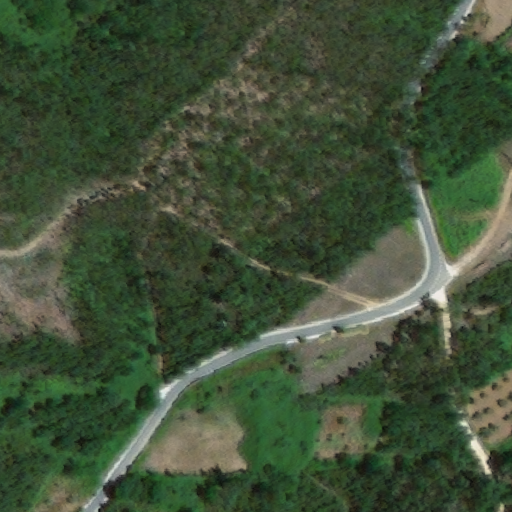}
      \end{minipage}
  }
  \subfigure[$\pi\mathbf{X}^{t_1}$]{
      \begin{minipage}[b]{0.3\linewidth}
          \includegraphics[width=\linewidth]{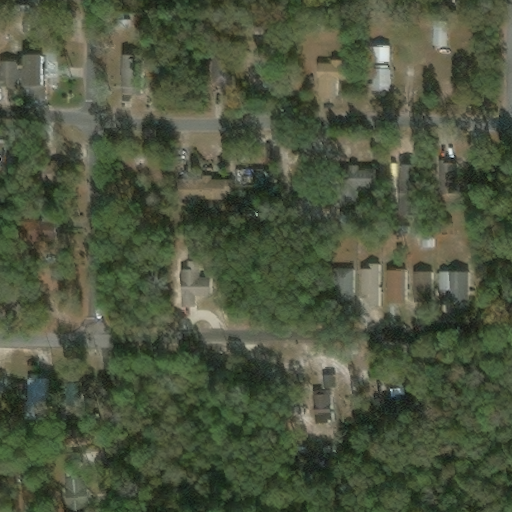}\vspace{4pt}
          \includegraphics[width=\linewidth]{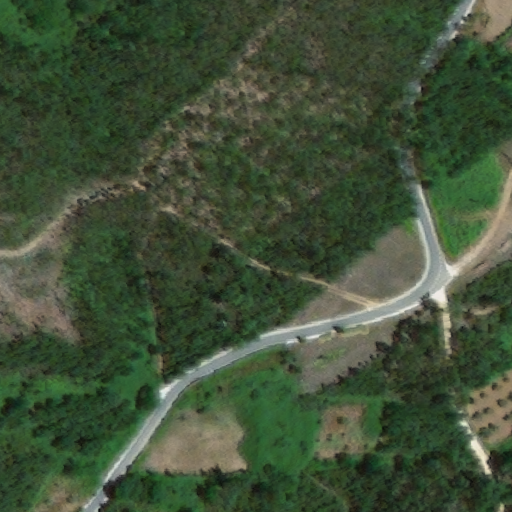}\vspace{4pt}
          \includegraphics[width=\linewidth]{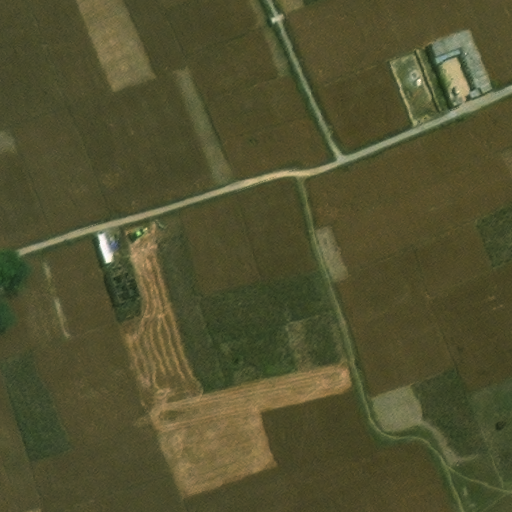}
      \end{minipage}
  }
  \subfigure[change label]{
      \begin{minipage}[b]{0.3\linewidth}
          \includegraphics[width=\linewidth]{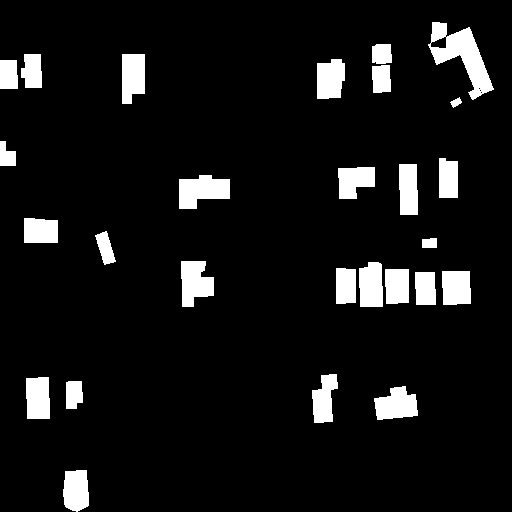}\vspace{4pt}
          \includegraphics[width=\linewidth]{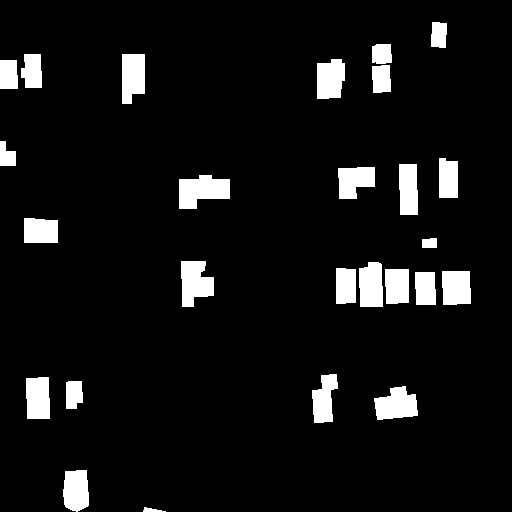}\vspace{4pt}
          \includegraphics[width=\linewidth]{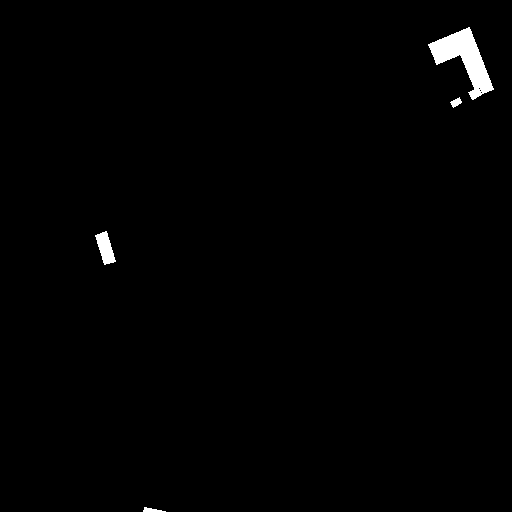}
      \end{minipage}
  }
  \caption{Pseudo bitemporal image pairs (a case of mini-batch of three images) for single-temporal supervised learning.
  $\mathbf{X}^{t_1}$, $\pi\mathbf{X}^{t_1}$ are the original sequence and the new sequence generated by a random permutation $\pi$.
  The change label is obtained by their semantic labels.
  }
  \label{fig:pseudo_bitemp}\vspace{-2mm}
\end{figure}

\begin{figure}
  \centering
  \begin{overpic}[width=\linewidth]{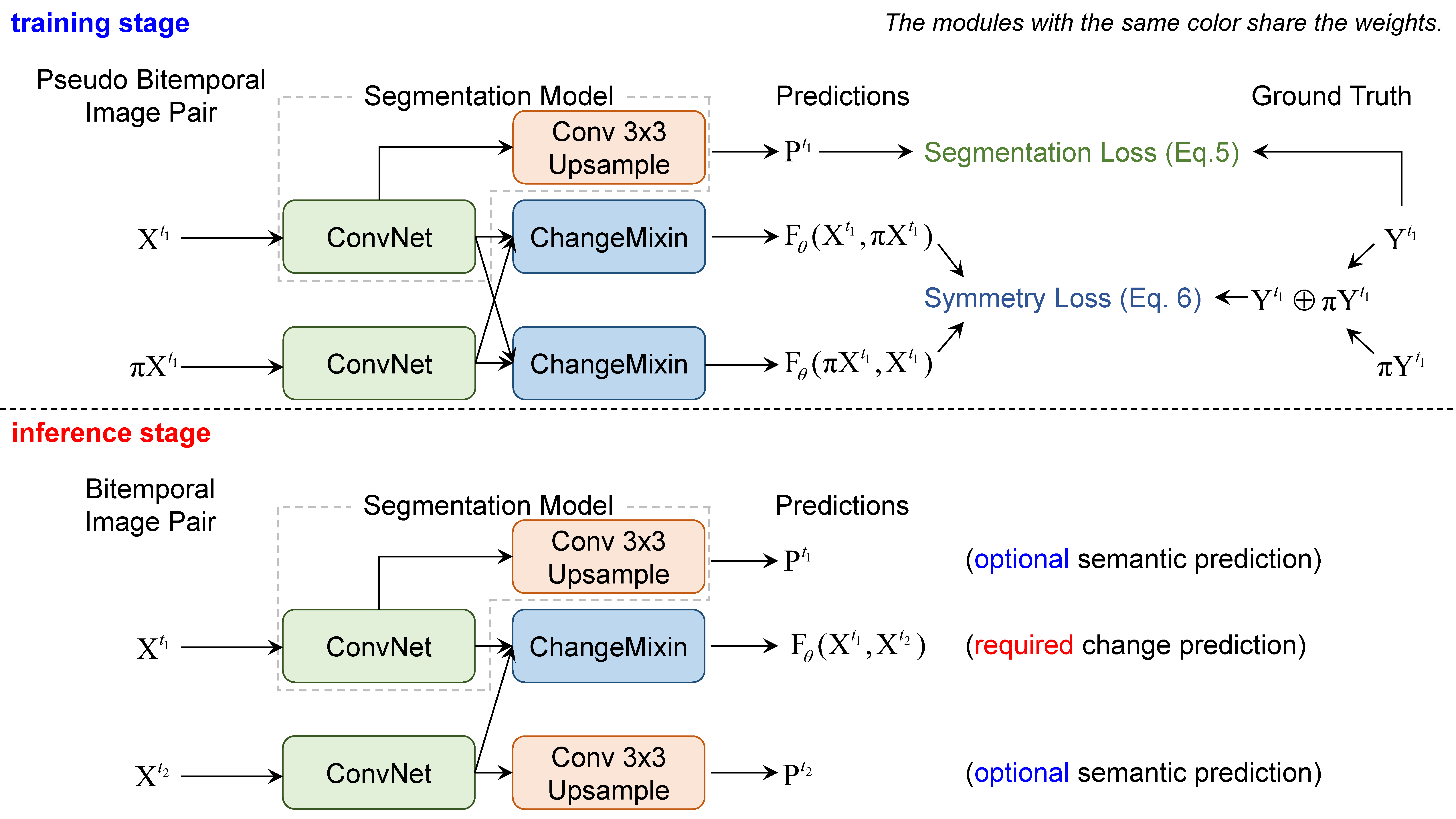}
     \end{overpic}
  \caption{\textbf{Overview of ChangeStar}. 
  The network architecture of ChangeStar is made up of an arbitrary deep semantic segmentation model and a ChangeMixin module.
  ChangeStar can be end-to-end trained by a segmentation loss and a symmetry loss with only single-temporal supervision.
  During training, weight sharing strategy is applied to the segmentation model and the ChangeMixin module.
  }
  \label{fig:overview}\vspace{-4mm}
\end{figure}

\subsection{Single-Temporal Supervised Learning}
The key idea of single-temporal supervised learning (STAR) is to learn a generalizable object change detector from arbitrary image pairs with only semantic labels via Eq.~\ref{eqn:gen}, as shown in Fig.~\ref{fig:overview}.
To provide change supervisory signals with single-temporal data, pseudo bitemporal image pairs are first constructed.
Leveraging pseudo bitemporal image pairs, bitemporal data in the original learning problem (Eq.~\ref{eqn:org}) can be replaced with single-temporal data, thus the learning problem can be reformulated as:
\begin{equation}
  \mathop{{\rm min}}\limits_{\theta} \, \mathcal{L}(\mathbf{F}_{\theta}(\mathbf{X}^{t_1}, \pi\mathbf{X}^{t_1}), \mathbf{Y}^{t_1} \oplus \pi\mathbf{Y}^{t_1})
\end{equation}
where pseudo bitemporal image pairs $\mathbf{X}^{t_1}, \pi\mathbf{X}^{t_1}$ with their change label $\mathbf{Y}^{t_1} \oplus \pi\mathbf{Y}^{t_1}$ provide single-temporal supervision.
The superscript $t_1$ is only used to represent that the data is single-temporal.

\subsubsection{Constructing Pseudo Bitemporal Image Pair}
To provide change supervisory signals with single-temporal data, we first construct pseudo bitemporal image pairs in a mini-batch and then assign labels to them during training.

\noindent  \textbf{Random Permutation in Mini-batch.}
Given a mini-batch single-temporal images $\mathbf{X}^{t_1}$ with its semantic labels $\mathbf{Y}^{t_1}$, $\mathbf{X}^{t_1}$ can be seen as a sequence $\{\mathbf{X}^{t_1}_1, ...,  \mathbf{X}^{t_1}_n\}$.
We use a random permutation $\pi \in S_n$ of this sequence to generate a new sequence $\pi\mathbf{X}^{t_1}$ to replace the $\mathbf{X}^{t_2}$, where $S_n$ denotes the all permutations of indices $\{1, ..., n\}$ except the permutations that cause any same element with the original sequence, and $\pi\mathbf{X}^{t_1}$ denotes the sequence $\{\mathbf{X}^{t_1}_{\pi(1)}, ...,  \mathbf{X}^{t_1}_{\pi(n)}\}$.
Fig.~\ref{fig:pseudo_bitemp} (a) and (b) present the original sequence of three images and the new sequence in case of a mini-batch of three images.

\noindent \textbf{Label Assignment.}
Different from manually pairwise dense labeling for bitemporal supervised learning, change labels are automatically generated by single-temporal semantic labels for STAR.
Without loss of generality, we discuss the label assignment for binary object change for simplicity. 
The positive labels of object change are assigned to the pixel positions in which the object of interest only once appeared.
If there are two object instances overlapped at pseudo bitemporal images, the pixel positions in the overlapping area are assigned as negative samples.
Because the object change is only semantic-aware, not instance-aware.
The rest of the pixel positions are assigned as negative samples.
To implement this label assignment, logical exclusive OR (\texttt{xor}) operation is a natural choice to obtain change labels with semantic labels $\mathbf{Y}^{t_1}$.
In this way, change labels $\mathbf{Y}^{t_1 \rightarrow t_2}$ in Eq.~\ref{eqn:org} can be replaced with $\mathbf{Y}^{t_1}\oplus\pi\mathbf{Y}^{t_1}$, where $\oplus$ denotes the \texttt{xor} operation, thus providing change supervisory signals with single-temporal data.
Fig.~\ref{fig:pseudo_bitemp} (c) demonstrates the generated change labels.

\subsubsection{Multi-task Supervision}
The overall objective function $\mathcal{L}$ is a multi-task objective function, which is used to sufficiently exploit single-temporal semantic labels for joint object segmentation and object change detection, as follows:
\begin{equation}
  \mathcal{L} = \mathcal{L}_{\texttt{seg}} + \mathcal{L}_{\texttt{change}}
\end{equation}
This work focus on the fundamental problem: binary object change, thus, there is only one type of object of interest.
Therefore, we introduce the objective functions for binary classification, whereas it is straightforward to extend this to the multi-class case.

\noindent \textbf{Semantic Supervision.}
For binary object segmentation, we adopt binary cross-entropy loss $\mathcal{L}_{\texttt{binary}}$ as $\mathcal{L}_{\texttt{seg}}$ to provide semantic supervision, as follows:
\begin{equation}
  \mathcal{L}_{\texttt{binary}}(p, y) = - [y{\rm log}(p) + (1-y){\rm log}(1-p)]
\end{equation}
where $y \in \{0, 1\}$ specifies the ground-truth class and $p \in [0, 1]$ denotes predicted probability for positive class.

\noindent \textbf{Change Supervision by Temporal Symmetry.}
Temporal symmetry is a mathematical property of binary object change, which indicates that binary object change is undirected, i.e. $Y^{t_1 \rightarrow t_2} = Y^{t_2 \rightarrow t_1}$.
Intuitively, the outputs of binary object change detector on the bitemporal image pair should follow this property.
This means that the binary object change detector should not fit the temporal direction under the constraint of temporal symmetry.
Motivated by this, we further propose symmetry loss for binary object change detection, which is formulated as follows:
\begin{equation}
      \begin{split}
          \mathcal{L}_{\texttt{change}} = & \frac{1}{2}[\mathcal{L}_{\texttt{binary}}(\mathbf{F}_\theta(\mathbf{X}^{t_1}, \pi \mathbf{X}^{t_1}), \mathbf{Y}^{t_1} \oplus \pi\mathbf{Y}^{t_1}) \\
           + &\mathcal{L}_{\texttt{binary}}(\mathbf{F}_\theta(\pi \mathbf{X}^{t_1}, \mathbf{X}^{t_1}), \mathbf{Y}^{t_1} \oplus \pi\mathbf{Y}^{t_1})] \\
      \end{split}
  \end{equation}
The symmetry loss features an inductive bias provided by temporal symmetry, which servers as a regularization term to alleviate the overfitting problem in binary object change detection.

\subsection{Network Architecture of ChangeStar}
\begin{figure}
  \centering
  \includegraphics[width=0.8\linewidth]{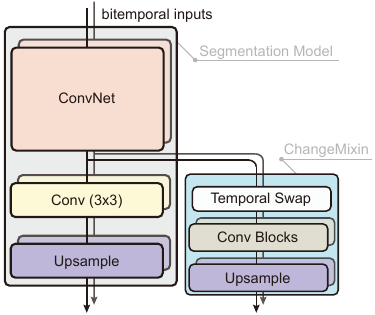}
  \caption{\textbf{Network architecture of ChangeStar}. 
  The network architecture of ChangeStar is made up of a deep segmentation model and a ChangeMixin module.
  The ChangeMixin module contains a temporal swap module and many conv layers, each followed by BN and ReLU.
  }
  \label{fig:arch}
\end{figure}

ChangeStar is a simple yet unified network composed of a deep semantic segmentation model and the ChangeMixin module.
This design is inspired by reusing the modern semantic segmentation architecture because semantic segmentation and object change detection are both dense prediction tasks.
To this end, we design the ChangeMixin module to enable any off-the-shelf deep semantic segmentation model to detect object change.
Fig.~\ref{fig:arch} presents the overall architecture of ChangeStar.

\noindent \textbf{Any Segmentation Model.}
The deep semantic segmentation model is used to extract a convolutional feature map for each image of the bitemporal inputs, respectively.
The top block of a segmentation model is always a 3$\times$3 conv layer with $C$ filters, followed by an upsampling layer, where $C$ is the number of classes and the upsampling scale is equal to the output stride of the specific segmentation model.
The feature map for object segmentation is computed by the whole segmentation model, while the feature map for object change detection is only computed by the ConvNet part of the segmentation model.

\noindent \textbf{ChangeMixin.}
The ChangeMixin module is composed of a temporal swap module (TSM) and a small FCN composed of $N$ 3$\times$3 conv layers, each with $d_c$ filters and each followed by BN and ReLU.
Besides, a bilinear upsampling layer followed by a sigmoid activation is attached to output the binary predictions per pixel.
The temporal swap module (Eq.~\ref{eqn:tsm}) is responsible for temporal symmetry, providing an inductive bias in the network architecture, which takes bitemporal feature maps as input and then concatenates them along the channel axis in two different temporal permutations.
\begin{equation}\label{eqn:tsm}
  \texttt{TSM}(\mathbf{T}_1, \mathbf{T}_2) = \texttt{cat}(\mathbf{T}_1, \mathbf{T}_2), \texttt{cat}(\mathbf{T}_2, \mathbf{T}_1)
\end{equation}
where $\mathbf{T}_1$ and  $\mathbf{T}_2$ denote bitemporal feature maps, respectively.
During training, the small FCN is attached to each output of TSM and the weight of the small FCN is shared.
During inference, the small FCN is only attached to the first output of TSM because we find that two outputs are temporal-symmetric in the converged model.
We use $N = 4$ and $d_c = 16$ for a better trade-off between speed and accuracy.

\section{Experiments}
\label{sec:exp}
We present experimental results on two HSR remote sensing building change detection datasets using the model trained on two HSR remote sensing building segmentation datasets with different domains, respectively, for a comprehensive analysis of the proposed method.
\subsection{Experimental Setting}

\noindent\textbf{Training Datasets.}
Two HSR remote sensing building segmentation datasets were used to train segmentation models and object change detectors by single-temporal supervision.
\begin{itemize}
    \vspace{-0.1in}
    \item \textbf{xView2 pre-disaster.} We used a subset of the xView2 dataset \cite{gupta2019creating}, namely xView2 pre-disaster, which is made up of the pre-disaster images and their annotations from \texttt{train} split and \texttt{tier3} split.
          The xView2 pre-disaster dataset consists of 9,168 HSR optical remote sensing images with a total of 316,114 building instances annotations in the context of the sudden-onset natural disaster.
          The images were collected from the Maxar / DigitalGlobe Open Data Program\footnote{https://www.digitalglobe.com/ecosystem/open-data}, and each image has a spatial size of 1,024$\times$1,024 pixels.
          \vspace{-0.1in}
    \item \textbf{SpaceNet2.}
          The public SpaceNet2 dataset \cite{van2018spacenet} consists of 10,590 HSR optical remote sensing images in the context of the urban scenarios, which were collected from DigitalGlobe's WorldView-3 satellite.
          This dataset also provides the annotation of 219,316 urban building instances.
          Each image has a spatial size of 650$\times$650 pixels with a spatial resolution of 0.3 m.
          In this study, we only used 3-bands pansharpened RGB images and their annotations.
\end{itemize}

\noindent\textbf{Evaluation Datasets.}
Two large scale HSR remote sensing building change detection datasets were used to evaluate the performance of object change detection.
\begin{itemize}
    \vspace{-0.05in}
    \item \textbf{WHU building change detection.} This dataset \cite{ji2018fully} consists of two aerial images obtained in 2012 and 2016 at same area of 20.5 km$^2$, which contains 12,796 and 16,077 building instances respectively.
          Each image has a spatial size of 15,354$\times$32,507 pixels with a spatial resolution of 0.2 m.
          There were a large number of rebuilt buildings and new constructions in this area because of a 6.3-magnitude earthquake in February 2011.
          \vspace{-0.05in}
    \item \textbf{LEVIR-CD.} LEVIR-CD dataset \cite{chen2020spatial} consists of 637 HSR bitemporal remote sensing image pairs, which were collected from Google Earth platform.
          Each image has a spatial size of 1024$\times$1024 pixels with a spatial resolution of 0.5 m.
          For annotation, this dataset provides a total of 31,333 change labels of building instances but without semantic labels.
          This dataset includes not only building appearing but also building disappearing for more general building changes.
          LEVIR-CD dataset is officially split into \texttt{train}, \texttt{val}, and \texttt{test}, three parts of which include 44,564, and 128 pairs, respectively.
          If not specified, the whole dataset (LEVIR-CD$^{\texttt{all}}$) is used for evaluation.
\end{itemize}

\begin{table*}[htb]
    \caption{Change detection IoU (\%) and F$_1$ (\%) on WHU building change detection and LEVIR-CD datasets.
        The backbone network is ResNet-50 for all models.
        All methods were trained using only single-temporal images and their semantic segmentation labels.
        \label{tab:benchmark}}
    \centering
    \renewcommand{\arraystretch}{1.5}
    \resizebox{\linewidth}{!}{
        \begin{tabular}{l|l|ll|ll|ll|ll|cc}
            \shline
            \multirow{3}{*}{Method} & \multirow{3}{*}{Segmentation Model}      & \multicolumn{4}{c|}{Train on xView2 pre-disaster} & \multicolumn{4}{c|}{Train on SpaceNet 2}       & \multirow{3}{*}{$\Delta$Params (M)} & \multirow{3}{*}{$\Delta$MAdds (B)}                                                                                                   \\ \cline{3-10}
                                    &                                          & \multicolumn{2}{c|}{WHU}                          & \multicolumn{2}{c|}{LEVIR-CD$^{\texttt{all}}$} & \multicolumn{2}{c|}{WHU}            & \multicolumn{2}{c|}{LEVIR-CD$^{\texttt{all}}$} &                 &                                                                   \\ \cline{3-10}
                                    &                                          & IoU (\%)                                          & F$_1$ (\%)                                     & IoU (\%)                            & F$_1$ (\%)                                     & IoU (\%)        & F$_1$ (\%)      & IoU (\%)        & F$_1$ (\%)      &      &      \\ \hline
            PCC                     & PSPNet \cite{zhao2017pyramid}            & 37.46                                             & 54.51                                          & 55.87                               & 71.69                                          & 21.39           & 35.25           & 10.19           & 18.50           & 0    & 0    \\
            ChangeStar (ours)       & + ChangeMixin                            & 56.44\up{18.98}                                   & 72.15\up{17.64}                                & 61.63\up{5.76}                      & 76.26\up{4.57}                                 & 25.56\up{4.17}  & 40.72\up{5.47}  & 15.25\up{5.06}  & 26.47\up{7.97}  & 0.16 & 0.63 \\ \hline
            PCC                     & DeepLab v3 \cite{chen2017rethinking}     & 32.46                                             & 49.01                                          & 54.77                               & 70.78                                          & 33.08           & 49.72           & 13.78           & 24.23           & 0    & 0    \\
            ChangeStar (ours)       & + ChangeMixin                            & 56.85\up{24.39}                                   & 72.49\up{23.48}                                & 60.94\up{6.17}                      & 75.73\up{4.95}                                 & 35.57\up{2.49}  & 52.48\up{2.76}  & 15.92\up{2.14}  & 27.46\up{3.23}  & 0.08 & 0.33 \\ \hline
            PCC                     & DeepLab v3+ \cite{chen2018encoder}       & 35.75                                             & 52.68                                          & 55.51                               & 71.38                                          & 23.90           & 38.58           & 9.80            & 17.85           & 0    & 0    \\
            ChangeStar (ours)       & + ChangeMixin                            & 52.01\up{16.26}                                   & 68.43\up{15.75}                                & 57.96\up{2.45}                      & 73.38\up{2.00}                                 & 38.42\up{15.42} & 55.51\up{16.93} & 22.22\up{12.42} & 36.36\up{18.51} & 0.08 & 0.33 \\ \hline
            PCC                     & Semantic FPN \cite{kirillov2019panoptic} & 38.66                                             & 55.76                                          & 56.19                               & 71.95                                          & 27.60           & 43.26           & 7.09            & 13.25           & 0    & 0    \\
            ChangeStar (ours)       & + ChangeMixin                            & 55.37\up{16.71}                                   & 71.27\up{15.51}                                & 65.21\up{9.02}                      & 78.94\up{6.99}                                 & 37.63\up{10.03} & 54.68\up{11.42} & 25.86\up{18.77} & 41.10\up{27.85} & 0.08 & 0.33 \\ \hline
            PCC                     & FarSeg \cite{zheng2020foreground}        & 31.66                                             & 48.09                                          & 55.09                               & 71.04                                          & 27.69           & 43.37           & 7.97            & 14.77           & 0    & 0    \\
            ChangeStar (ours)       & + ChangeMixin                            & 58.22\up{26.56}                                   & 73.59\up{25.50}                                & 65.71\up{10.62}                     & 79.31\up{8.27}                                 & 39.02\up{11.33} & 56.14\up{12.77} & 30.42\up{22.45} & 46.65\up{31.88} & 0.08 & 0.33 \\ \shline
        \end{tabular}
    }
    \vspace{-4mm}
\end{table*}

\vspace{-0.05in}
\noindent\textbf{Implementation detail.}
Unless otherwise specified, all models were trained for 40k iterations with a \texttt{poly} learning rate policy, where the initial learning rate was set to 0.03 and multiplied by $(1 - \frac{\texttt{step}}{\texttt{max\_step}})^{\gamma}$ with $\gamma = 0.9$.
We used SGD as the optimizer on single Titan RTX GPU with a mini-batch of 16 images, weight decay of 0.0001 and momentum of 0.9.
For training data augmentation, after horizontal and vertical flip, rotation of $90\cdot k~(k=1,2,3)$ degree, and scale jitter, the images are then randomly cropped into 512$\times$512 pixels for xView2 pre-disaster dataset and 256$\times$256 pixels for SpaceNet2 dataset.

\noindent\textbf{Metrics.}
The binary object change detection belongs to pixel-wise binary classification task, therefore we adopt intersection over union (IoU) and F$_1$ score to evaluate the object change detection.

\subsection{Main Results}
In the weakly-supervised setting that only single-temporal supervision is available, PCC series are reasonable baselines when using strong semantic segmentation models.
Thus, we compare ChangeStar against PCCs with many representative segmentation models \cite{zhao2017pyramid,chen2017rethinking,chen2018encoder,kirillov2019panoptic,zheng2020foreground}.
The results listed in Table~\ref{tab:benchmark} show that ChangeStar significantly outperforms PCC with different segmentation models in this challenging cross-domain evaluation.
Notably, these improvements only come at the cost of much slight overhead, which confirms the significance of learning object change representation.
Overall, training on the xView2 pre-disaster is obviously superior to training on SpaceNet2.
We conjecture that richer background distribution of the xView2 pre-disaster can provide more diverse positive samples, which facilitates more generalized object change representation.
Besides, the images of xView2 pre-disaster have multiple spatial resolution, while th images of SpaceNet2 have a fixed spatial resolution of 0.3 m.

\subsection{Ablation Study}
To delve into the proposed method, we conducted comprehensive experiments using ChangeStar based on FarSeg with ResNet-50 if not specified, since it is more robust than other variants of ChangeStar.

\begin{figure}[htb]
    \centering
    \subfigure[Number of layers]{
        \begin{minipage}[b]{0.49\linewidth}
            \includegraphics[width=\linewidth]{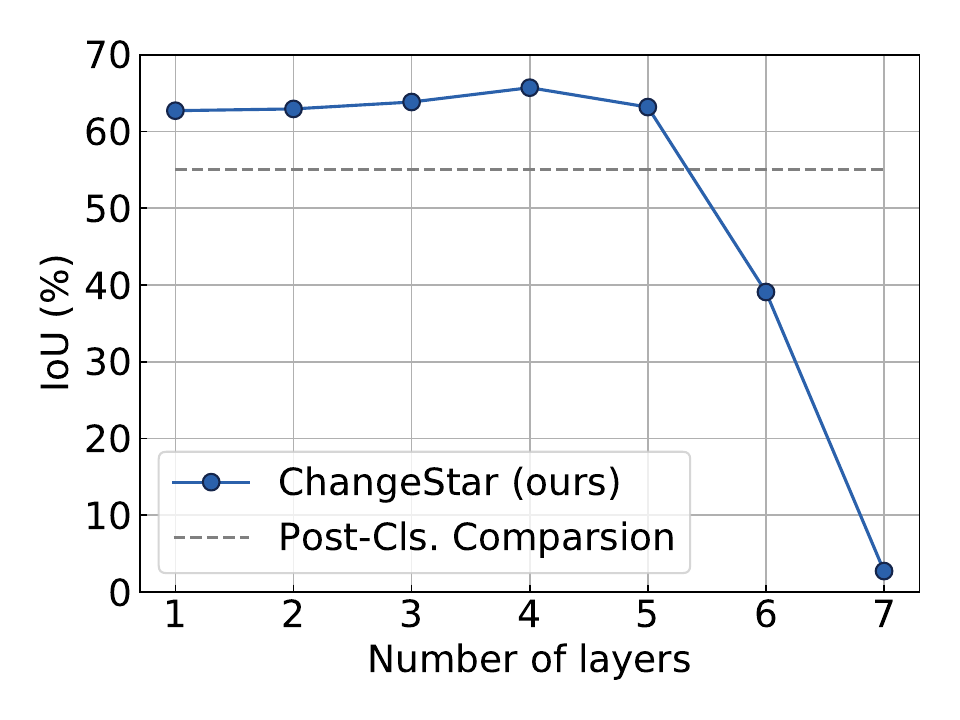}
        \end{minipage}
        \begin{minipage}[b]{0.49\linewidth}
            \includegraphics[width=\linewidth]{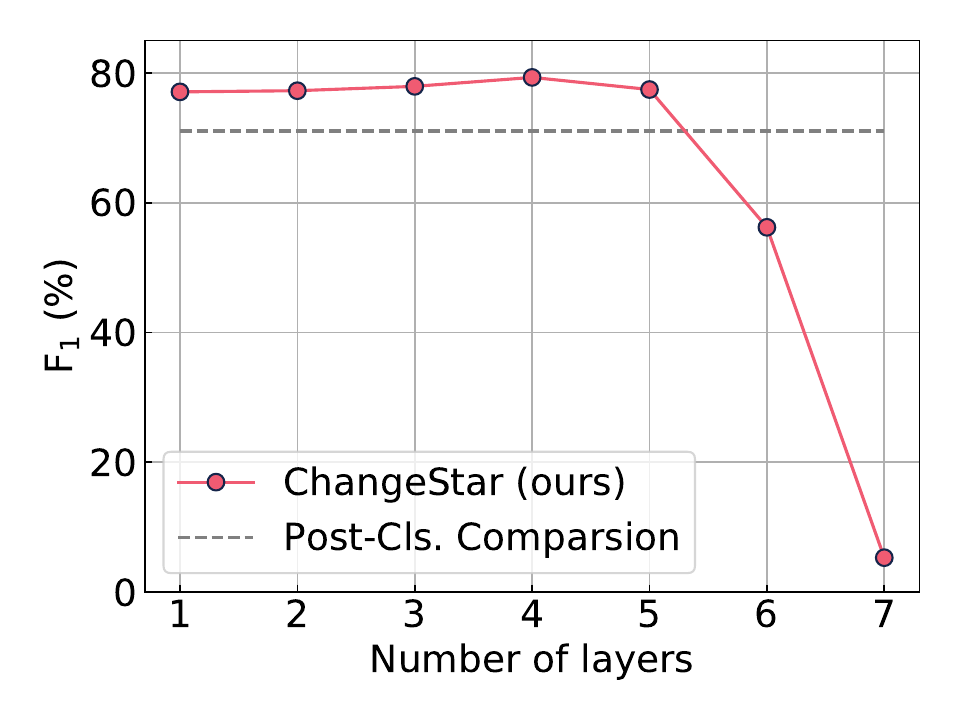}
        \end{minipage}
    }\vspace{-4mm}
    \subfigure[Number of channels]{
        \begin{minipage}[b]{0.49\linewidth}
            \includegraphics[width=\linewidth]{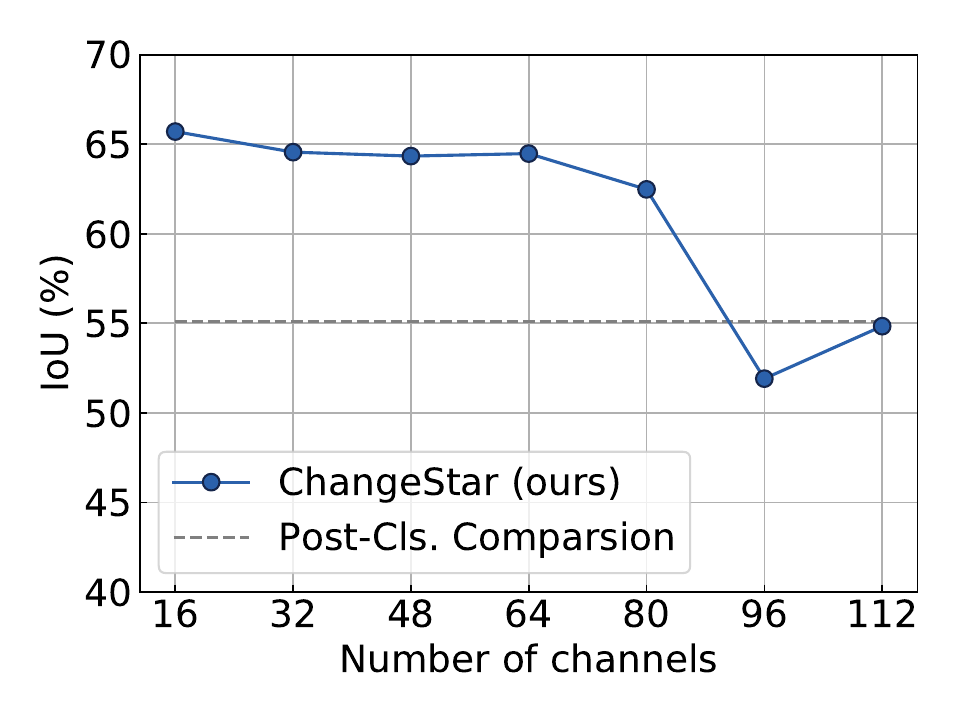}
        \end{minipage}
        \begin{minipage}[b]{0.49\linewidth}
            \includegraphics[width=\linewidth]{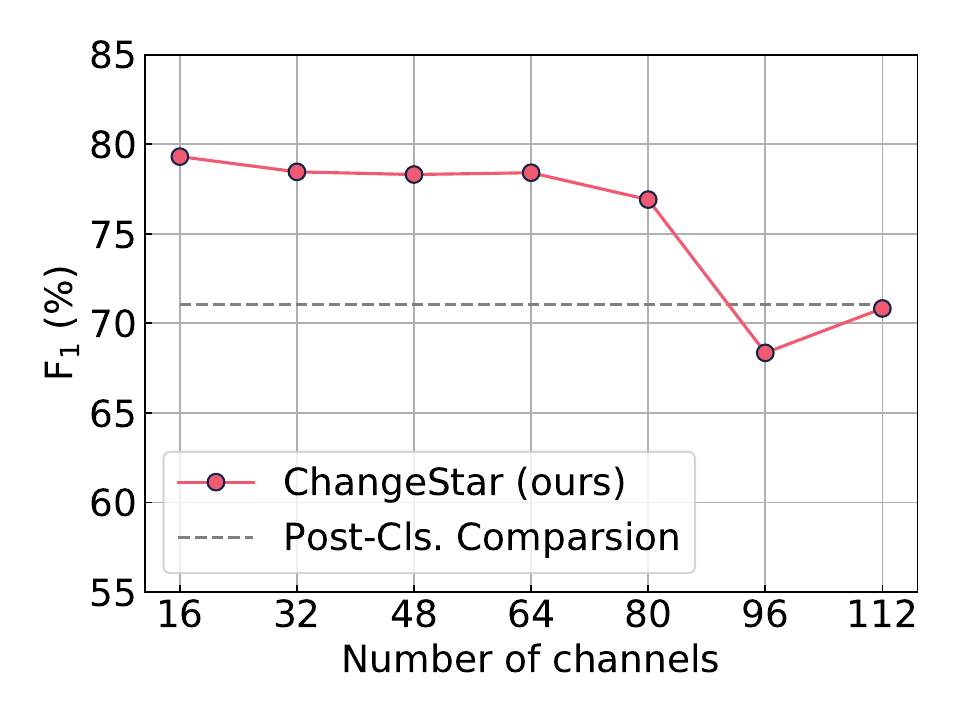}
        \end{minipage}
    }
    \vspace{-4mm}
    \caption{Object change detection results on LEVIR-CD$^{\texttt{all}}$ using different hyperparameter settings of the ChangeMixin Module.
    }
    \label{fig:abs_arch}
    \vspace{-4mm}
\end{figure}

\noindent\textbf{Architecture of ChangeMixin.}
The ChangeMixin module is the most important component in ChangeStar, which introduces two hyper-parameters: $N$ (number of conv layers) and $d_c$ (number of convolutional filters).
The performance of ChangeStar with varying $N$ are presented in Fig.~\ref{fig:abs_arch} (a).
It can be found that over-deep convolutional subnetwork is harmful to the object change detection performance.
ChangeStar performs better than the post-classification comparison when $N<6$ and achieves best performance at $N = 4$.
The performance of ChangeStar with varying $d_c$ are presented in Fig.~\ref{fig:abs_arch} (b).
As $d_c$ increases, the performance constantly decreases and worse than the post-classification comparison when $d_c \ge 80$.
For a better trade-off between speed and accuracy, we use $N = 4$ and $d_c = 16$ as the default setting.

\vspace{-4mm}

\begin{table}[htb]
    \caption{Object change detection results on LEVIR-CD$^{\texttt{all}}$ for understanding the contribution of each component.
        \label{tab:as_component}}
    \centering
    \renewcommand{\arraystretch}{1.5}
    \resizebox{\linewidth}{!}{
        \begin{tabular}{l|ccc|cc}
            \shline
            Method                        & STAR         & Semantic Sup. & Temporal Sym. & IoU (\%) & F$_1$ (\%) \\ \shline
            (a) PCC                       &              & $\checkmark$  &               & 55.09    & 71.04      \\ \hline
            (b) Baseline                  & $\checkmark$ &               &               & 61.85    & 76.43      \\
            (c) Baseline w/ Semantic Sup. & $\checkmark$ & $\checkmark$  &               & 62.42    & 76.86      \\
            (d) Baseline w/ Temporal Sym. & $\checkmark$ &               & $\checkmark$  & 64.10    & 78.12      \\
            (e) ChangeStar                & $\checkmark$ & $\checkmark$  & $\checkmark$  & 65.71    & 79.31      \\
            \shline
        \end{tabular}
    }\vspace{-4mm}
\end{table}

\noindent\textbf{Importance of Semantic Supervision.}
Semantic supervision not only enables ChangeStar to segment objects but also can facilitate object change representation learning.
Table~\ref{tab:as_component} (b)/(c) and (d)/(e) show that the introduction of semantic supervision is positive for object change detection.
Quantitatively, semantic supervision improves the baseline by 0.57\% IoU and 0.43\% F$_1$ and improves the baseline with temporal symmetry by 1.61\% IoU and 1.19\% F$_1$.
This indicates that semantic representation provided by semantic supervision facilitates object change representation learning, and object change representation is stronger when possessing temporal symmetry.

\noindent\textbf{Importance of Temporal Symmetry.}
Temporal symmetry, as a mathematical property of binary object change, can provide a prior as regularization to learn more robust object change representation.
Table~\ref{tab:as_component} (a)/(d) and (c)/(e) shows that using temporal symmetry gives a 2.25\% IoU and 1.69\% F$_1$ gains over the baseline and gives a 3.29\% IoU and 2.45\% F$_1$ over the baseline with semantic supervision.
This indicates that it is significantly important to guarantee temporal symmetry in binary object change detection for STAR.
We can also find that temporal symmetry makes it better to learn object change representation from semantic representation.

\noindent\textbf{Label assignment.}
Here we discuss the impact of different label assignment strategies on accuracy. 
Table~\ref{tab:label_assign} presents that using \texttt{or} achieves 43.84\% IoU, while using \texttt{xor} achieves 65.71\% IoU. 
This is because these negative samples (i.e. overlapped region) are necessary to make the model learn to suppress false positives that occurred on objects that have not changed in the period from $t_1$ to $t_2$, which can be satisfied by \texttt{xor}.
However, \texttt{or} operation wrongly assigns their labels.

\vspace{-2mm}

\begin{table}[htb]
    \caption{The accuracy of different label assignment strategies.
        \label{tab:label_assign}}
    \centering
    \renewcommand{\arraystretch}{1.2}
    \begin{tabular}{l|cc}
        \shline
        Method       & IoU (\%) & F$_1$ (\%) \\ \hline
        \texttt{or}  & 43.84    & 60.96      \\
        \texttt{xor} & 65.71    & 79.31      \\
        \shline
    \end{tabular}
    \vspace{-2mm}
\end{table}

\begin{figure*}[htb]
    \centering
    \subfigure[$t_1$ Image]{
        \begin{minipage}[b]{0.15\linewidth}
            \includegraphics[width=\linewidth]{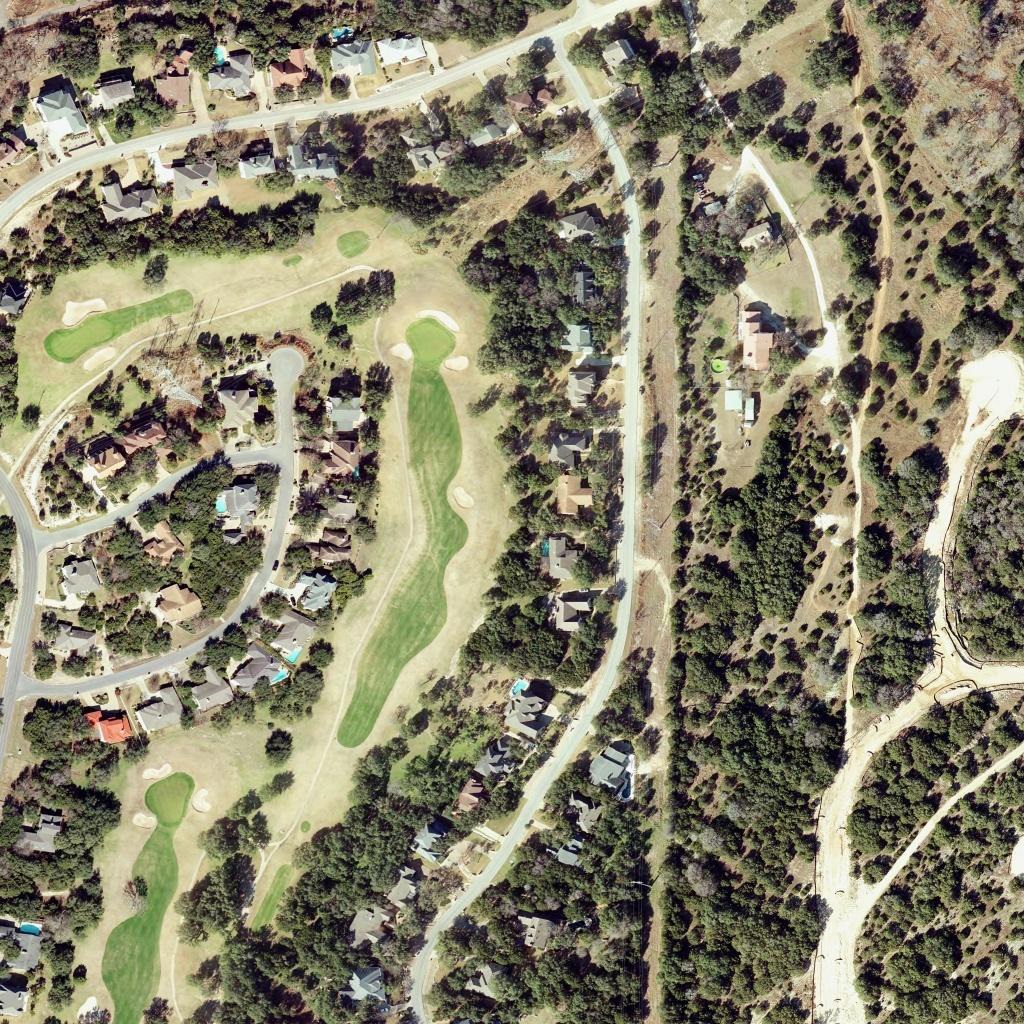}\vspace{4pt}
            \includegraphics[width=\linewidth]{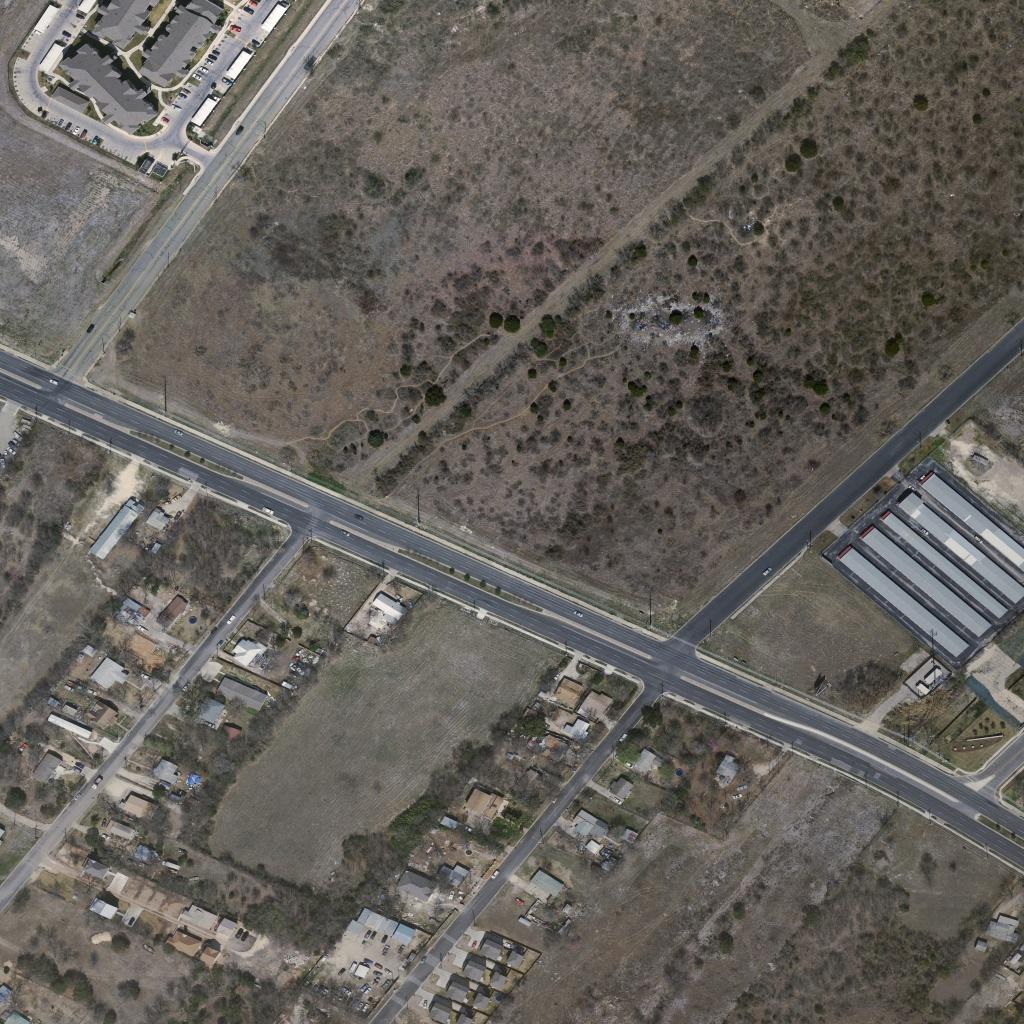}
        \end{minipage}
    }
    \subfigure[$t_2$ Image]{
        \begin{minipage}[b]{0.15\linewidth}
            \includegraphics[width=\linewidth]{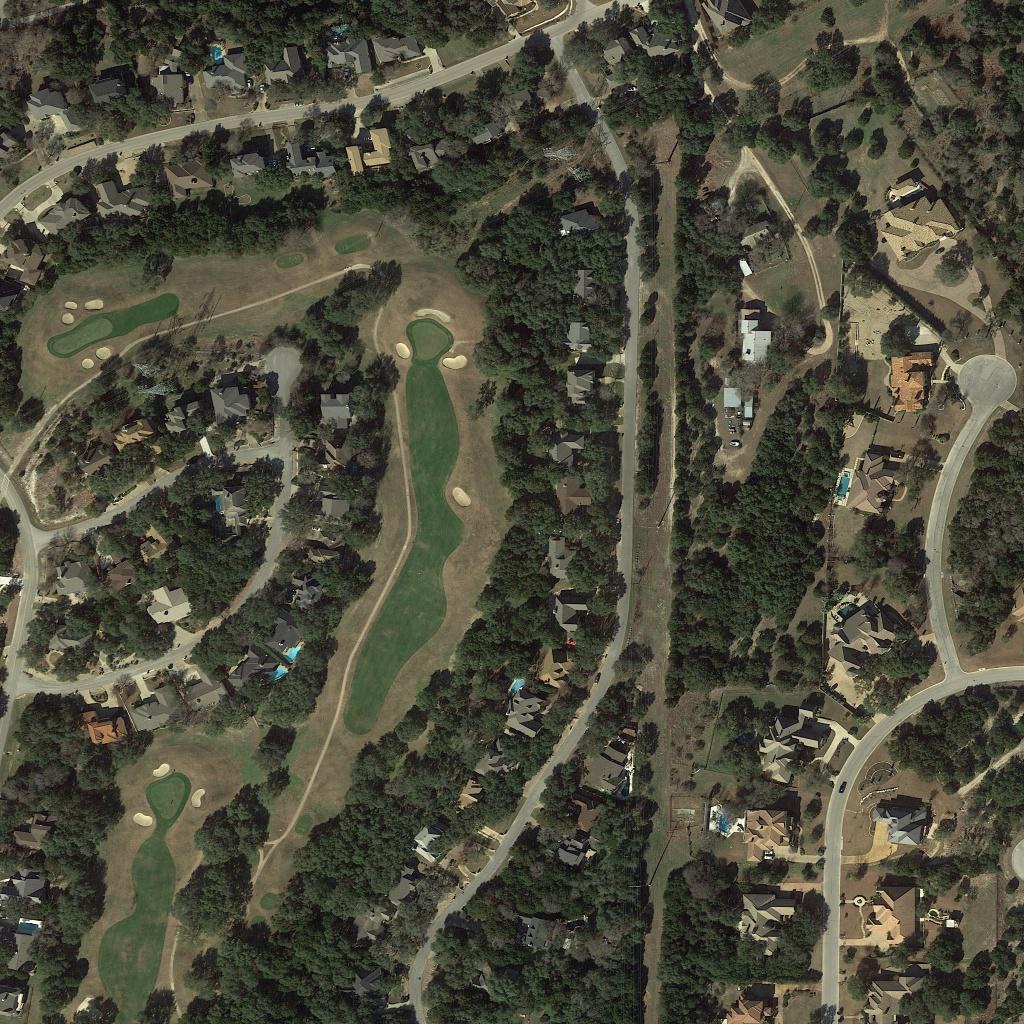}\vspace{4pt}
            \includegraphics[width=\linewidth]{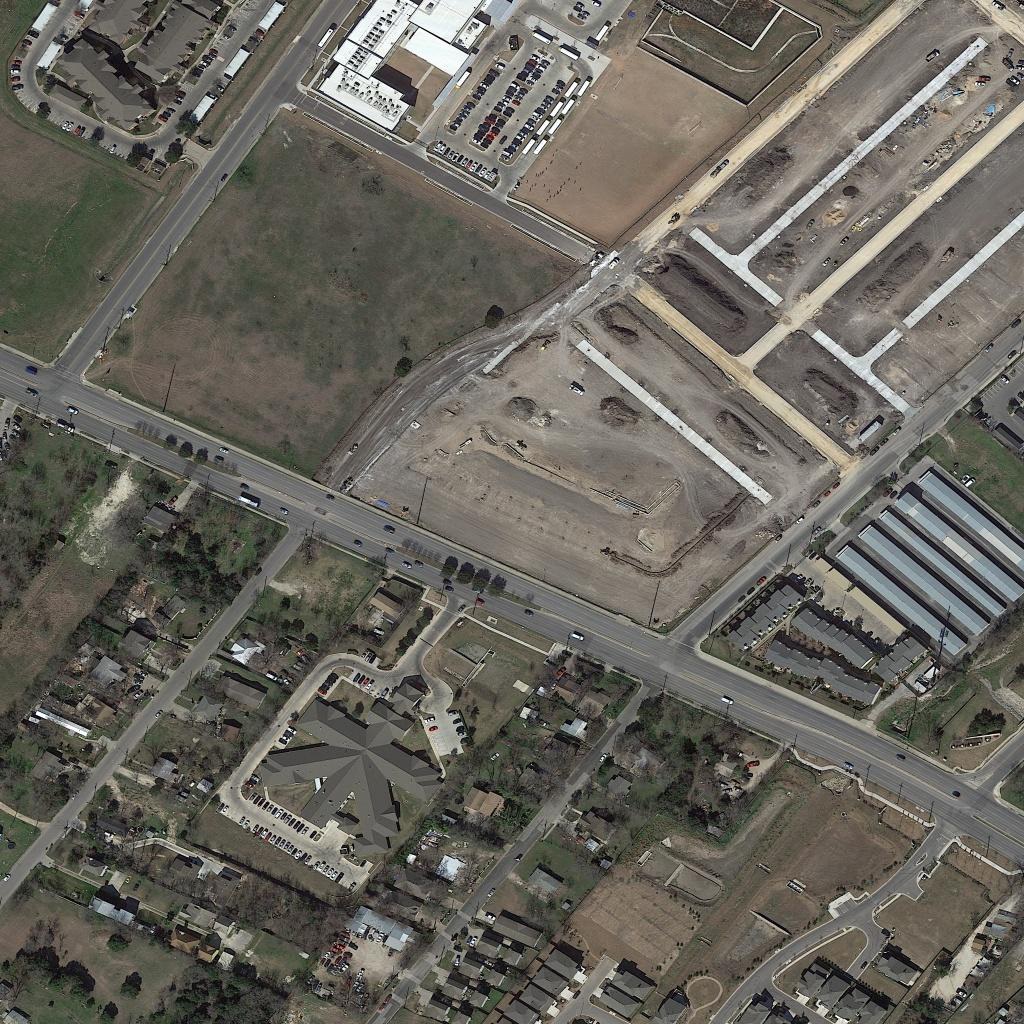}
        \end{minipage}
    }
    \subfigure[Ground Truth]{
        \begin{minipage}[b]{0.15\linewidth}
            \includegraphics[width=\linewidth]{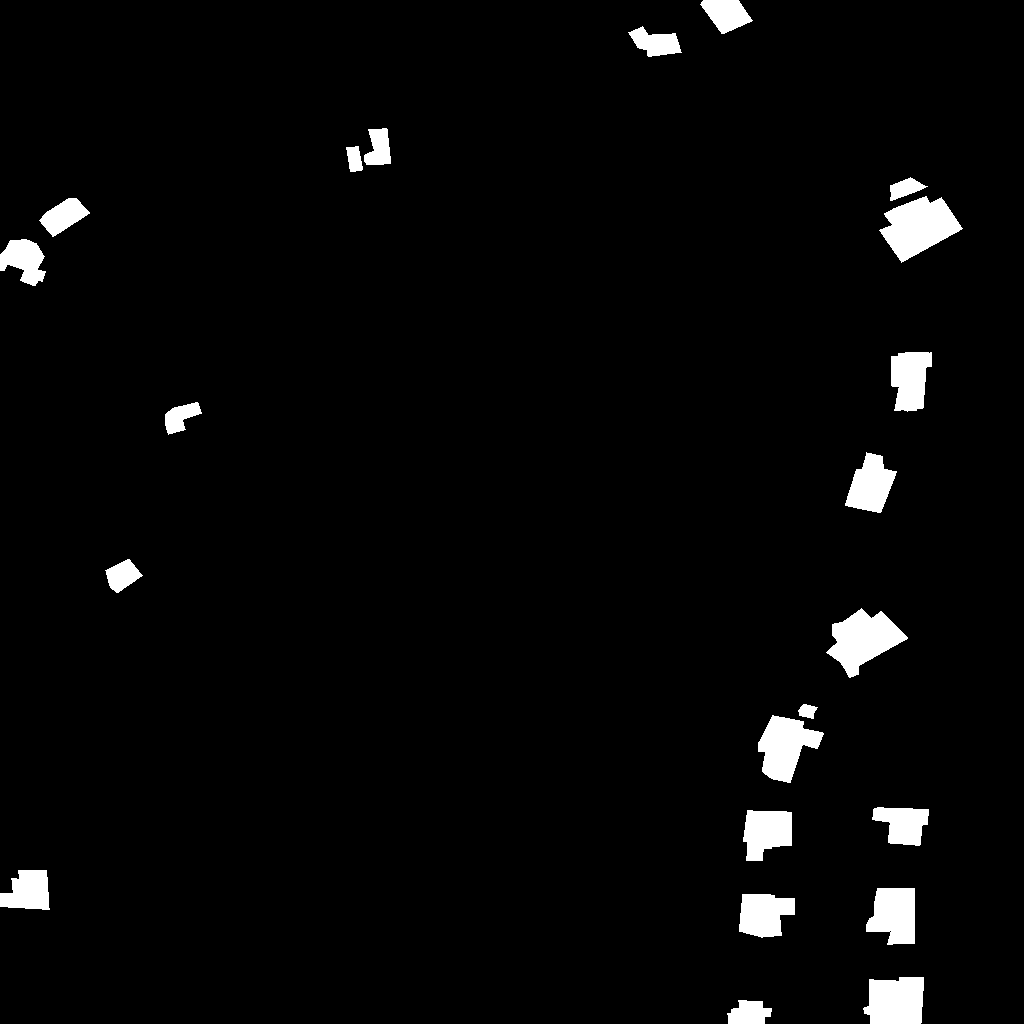}\vspace{4pt}
            \includegraphics[width=\linewidth]{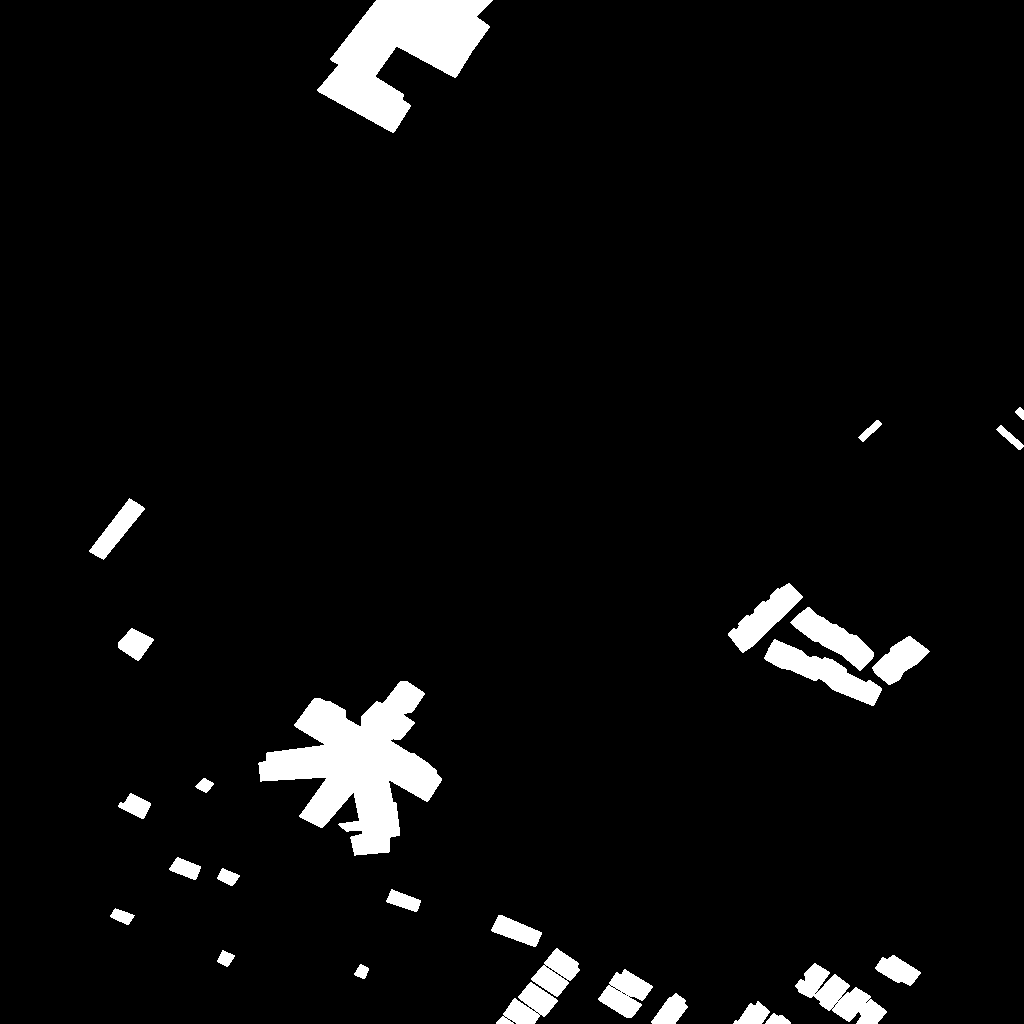}
        \end{minipage}
    }
    \subfigure[Bitemporal Sup.]{
        \begin{minipage}[b]{0.15\linewidth}
            \includegraphics[width=\linewidth]{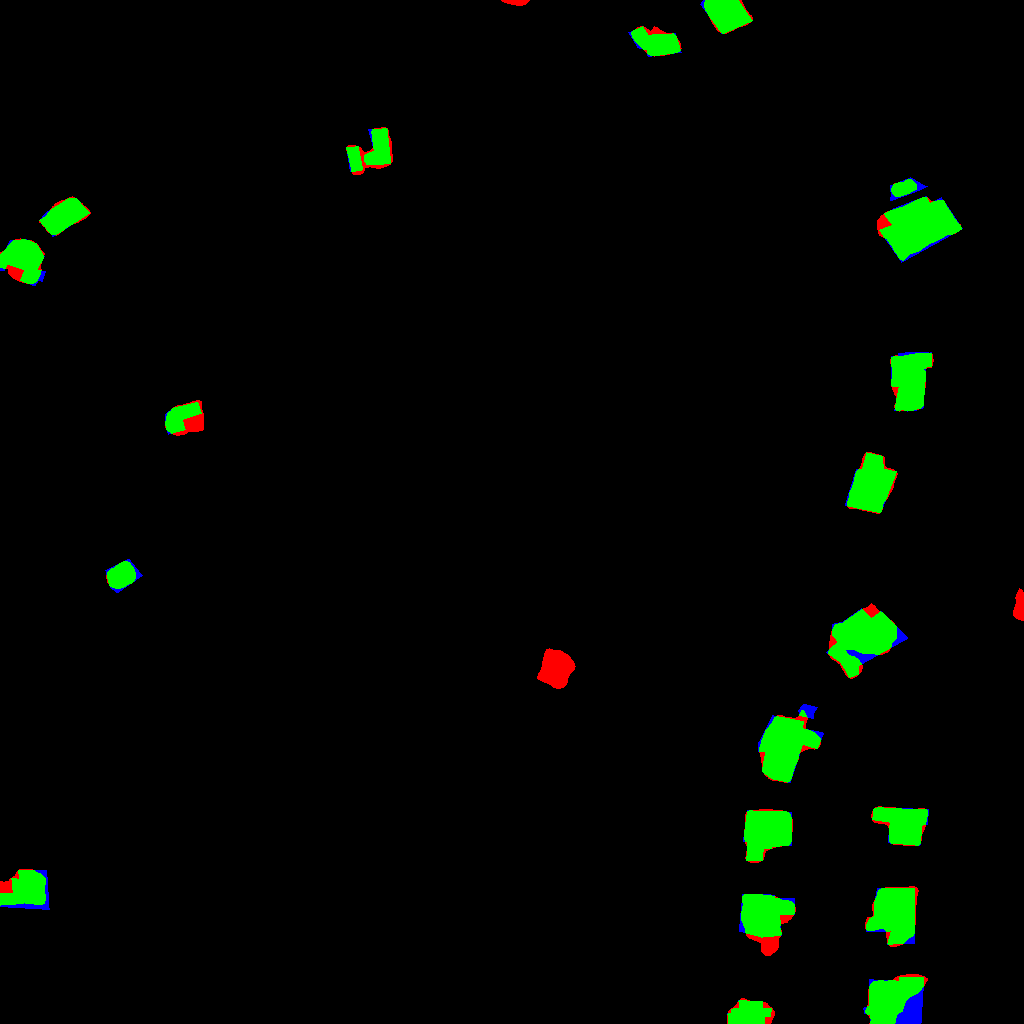}\vspace{4pt}
            \includegraphics[width=\linewidth]{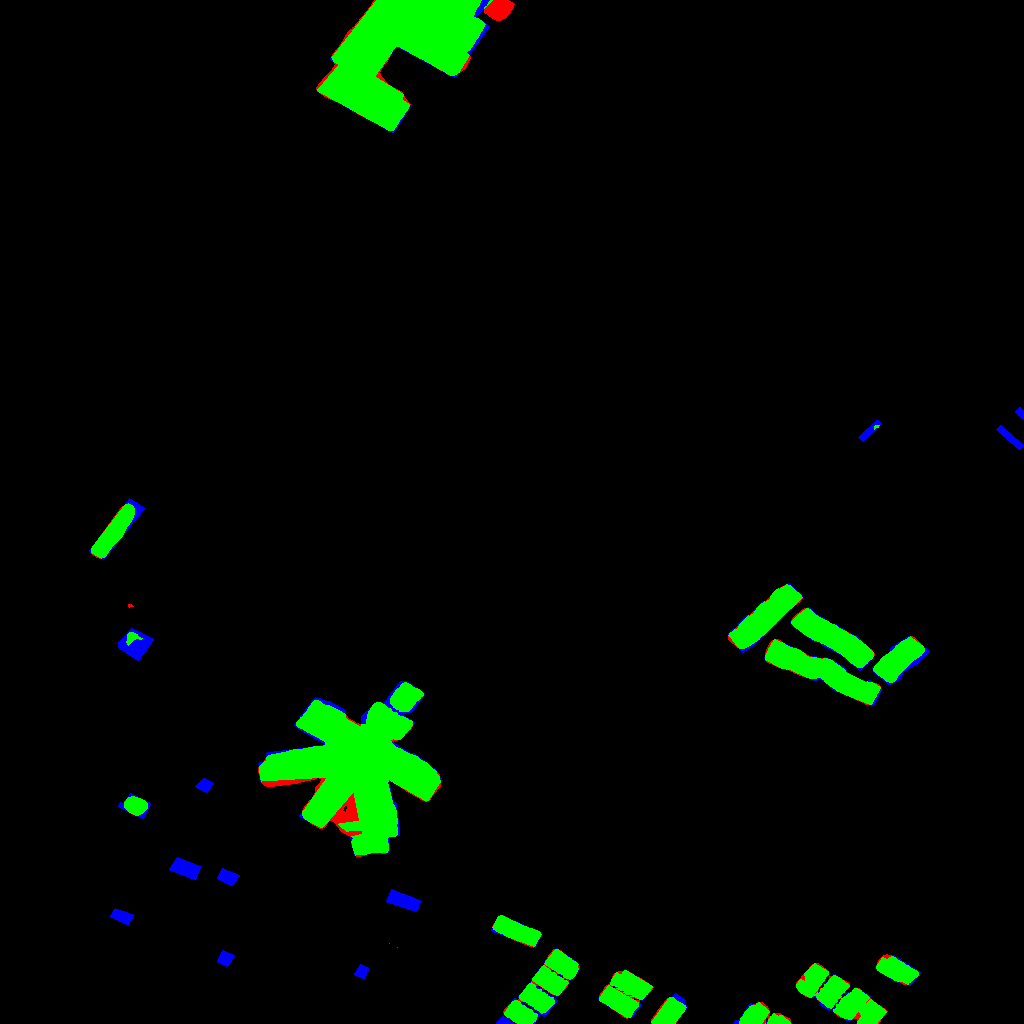}
        \end{minipage}
    }
    \subfigure[PCC]{
        \begin{minipage}[b]{0.15\linewidth}
            \includegraphics[width=\linewidth]{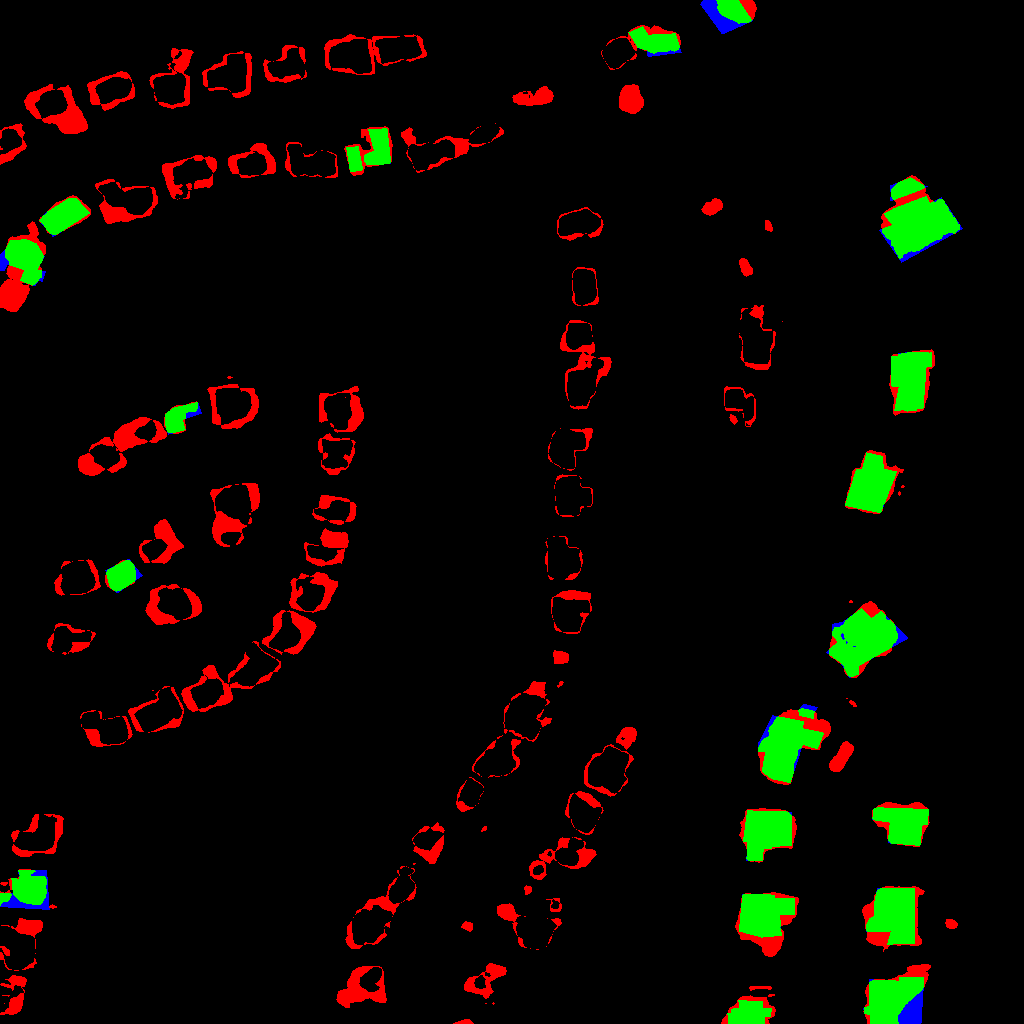}\vspace{4pt}
            \includegraphics[width=\linewidth]{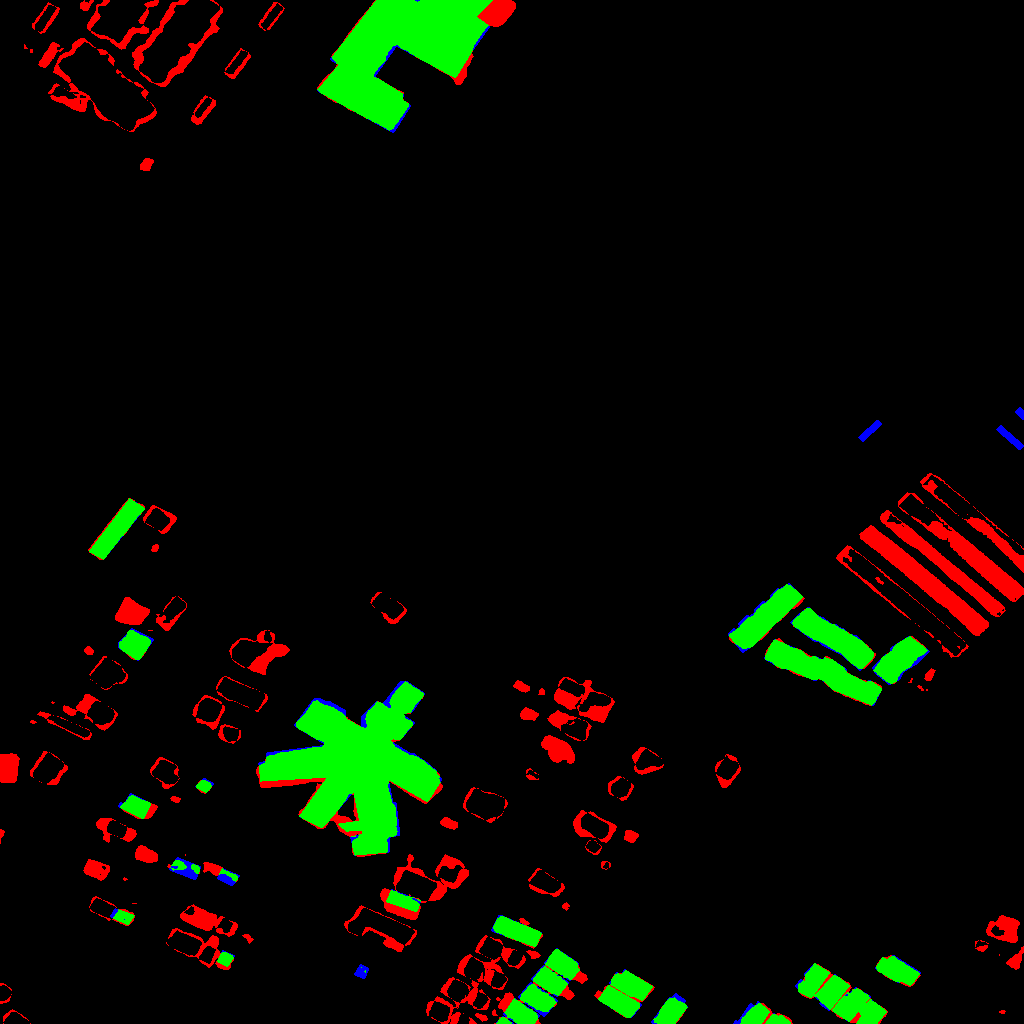}
        \end{minipage}
    }
    \subfigure[STAR]{
        \begin{minipage}[b]{0.15\linewidth}
            \includegraphics[width=\linewidth]{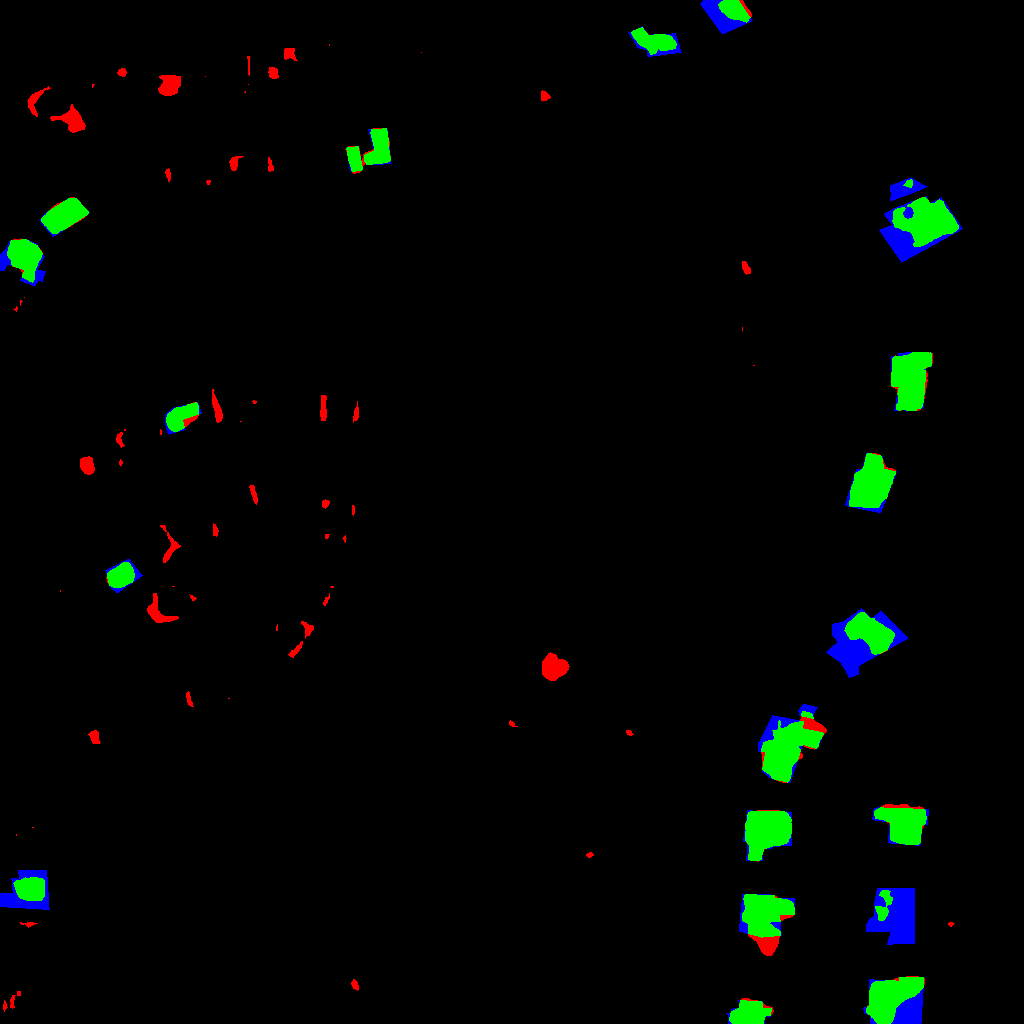}\vspace{4pt}
            \includegraphics[width=\linewidth]{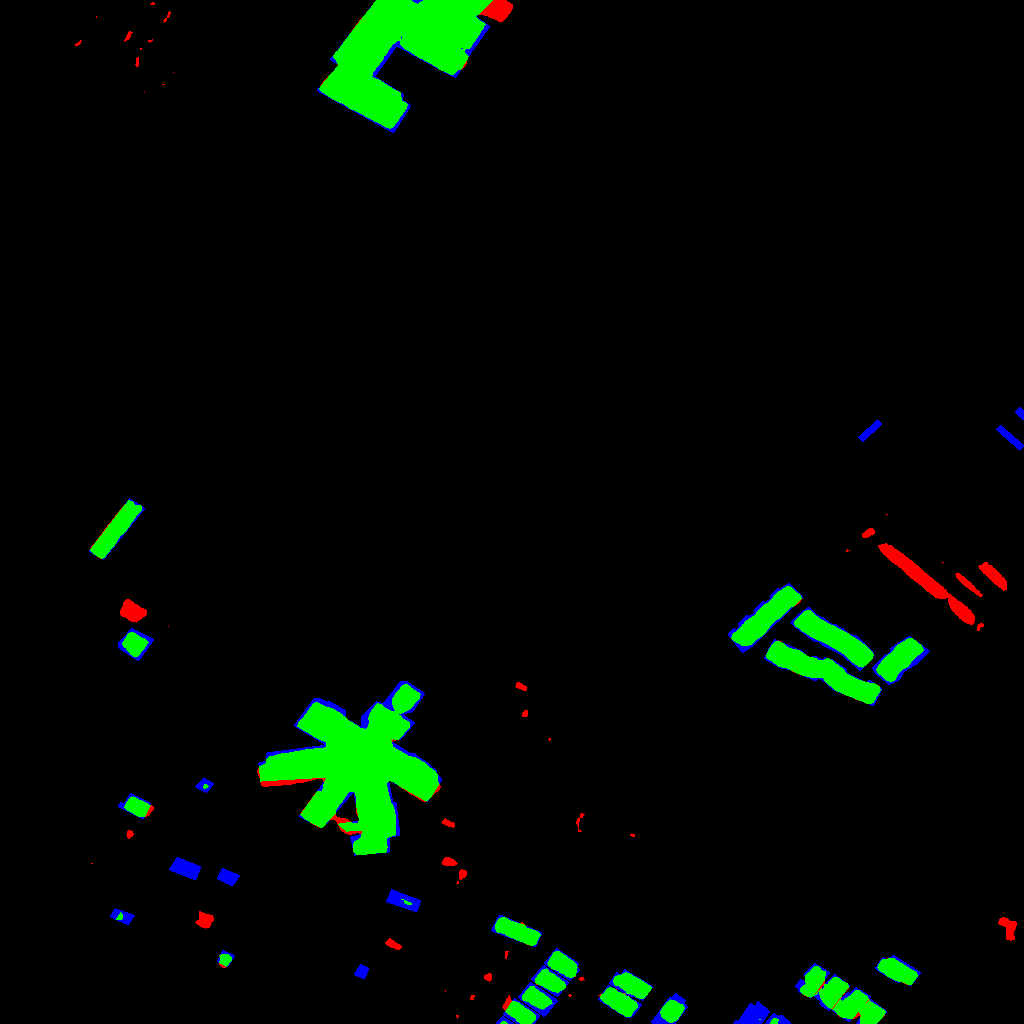}
        \end{minipage}
    }
    \vspace{-0.1in}
    \caption{Error analysis for ChangeStar with bitemporal supervision, PCC and ChangeStar with STAR.
        The basic segmentation model is FarSeg with ResNeXt-101 32x4d.
        The rendered colors represent \textcolor[rgb]{0,1,0}{true positives (TP)}, \textcolor[rgb]{1,0,0}{false positives (FP)}, and \textcolor[rgb]{0,0,1}{false negatives (FN)}.
    }
    \label{fig:error_maps}
    \vspace{-4mm}
\end{figure*}

\noindent\textbf{ChangeStar using Bitemporal Sup.}
ChangeStar is a object change detection architecture driven by STAR as default, but it also can be driven by bitemporal supervision.
We benchmark many variants of ChangeStar and the results are presented in Table~\ref{tab:bisup}.
\begin{table}[htb]
    \caption{Bitemporal supervised benchmark.
        All methods were trained on LEVIR-CD$^{\texttt{train}}$ and evaluated on LEVIR-CD$^{\texttt{test}}$ for fair comparison.
        \label{tab:bisup}}
    \centering
    \renewcommand{\arraystretch}{1.5}
    \resizebox{\linewidth}{!}{
        \begin{tabular}{l|l|l|ccc}
            \shline
            Method                                  & Backbone          & IoU (\%) & F$_1$ (\%) \\ \hline

            FCN + BAM \cite{chen2020spatial}        & ResNet-18         & -        & 85.7       \\
            FCN + PAM \cite{chen2020spatial}        & ResNet-18         & -        & 87.3       \\ \hline
            ChangeStar (PSPNet + ChangeMixin)       & ResNet-18         & 78.08    & 87.69      \\
            ChangeStar (DeepLab v3 + ChangeMixin)   & ResNet-18         & 77.95    & 87.61      \\
            ChangeStar (DeepLab v3+ + ChangeMixin)  & ResNet-18         & 81.32    & 89.70      \\
            ChangeStar (Semantic FPN + ChangeMixin) & ResNet-18         & 82.51    & 90.41      \\
            ChangeStar (FarSeg + ChangeMixin)       & ResNet-18         & 82.31    & 90.29      \\
            ChangeStar (FarSeg + ChangeMixin)       & ResNet-50         & 83.19    & 90.82      \\
            ChangeStar (FarSeg + ChangeMixin)       & ResNeXt-101 32x4d & 83.92    & 91.25      \\
            \shline
        \end{tabular}
    }
    \vspace{-2mm}
\end{table}
\begin{table}[htb]
    \caption{Bitemporal supervision versus single-temporal supervision.
        All methods were evaluated on LEVIR-CD$^{\texttt{test}}$ for consistent comparison.
        \label{tab:as_bi_st}}
    \centering
    \renewcommand{\arraystretch}{1.5}
    \resizebox{\linewidth}{!}{
        \begin{tabular}{l|l|l|ccc}
            \shline
            Method                              & Backbone          & Training data               & IoU (\%) & F$_1$ (\%) & F$_1$ gap (\%) \\ \hline
            \textit{Bitemporal Supervised}      &                   &                             &          &            &                \\
            ChangeStar (FarSeg + ChangeMixin)   & ResNet-18         & LEVIR-CD$^{\texttt{train}}$ & 82.31    & 90.29      & -              \\
            ChangeStar (FarSeg + ChangeMixin)   & ResNet-50         & LEVIR-CD$^{\texttt{train}}$ & 83.19    & 90.82      & -              \\
            ChangeStar (FarSeg + ChangeMixin)   & ResNeXt-101 32x4d & LEVIR-CD$^{\texttt{train}}$ & 83.92    & 91.25      & -              \\ \shline
            \textit{Single-Temporal Supervised} &                   &                             &          &            &                \\
            PCC (FarSeg)                        & ResNet-18         & xView2 pre-disaster         & 56.65    & 72.32      & -17.97         \\
            PCC (FarSeg)                        & ResNet-50         & xView2 pre-disaster         & 55.89    & 71.71      & -19.11         \\
            PCC (FarSeg)                        & ResNeXt-101 32x4d & xView2 pre-disaster         & 59.54    & 74.64      & -16.61         \\ \hline
            ChangeStar (FarSeg + ChangeMixin)   & ResNet-18         & xView2 pre-disaster         & 63.25    & 77.49      & -12.08         \\
            ChangeStar (FarSeg + ChangeMixin)   & ResNet-50         & xView2 pre-disaster         & 66.99    & 80.23      & -10.58         \\
            ChangeStar (FarSeg + ChangeMixin)   & ResNeXt-101 32x4d & xView2 pre-disaster         & 68.84    & 81.54      & -9.71          \\ \shline
        \end{tabular}
    }
    \vspace{-2mm}
\end{table}
We can find that atrous convolution based ChangeStars (PSPNet, DeepLab v3) achieves compatible results with spatial-temporal attention based methods (FCN + BAM and FCN + PAM).
When introducing encoder-decoder architecture, ChangeStars (DeepLab v3+, semantic FPN, FarSeg) achieves better performance by a large margin.
When further introducing FPN-family decoder, ChangeStars (semantic FPN, FarSeg) are superior to other variants.
We thus conclude that encoder-decoder and FPN architectures are more friendly to object change detection, which may attribute to the multi-scale problem \cite{zheng2020foreground}.
Besides, the deeper backbone network brings more accuracy gains, which achieves agreement to other vision tasks \cite{ren2015faster, he2016deep, he2017mask, tran2018closer}.
These results suggest that ChangeStar is a simple yet effective object change detection architecture.

\noindent\textbf{Bitemporal Sup. vs. Single-Temporal Sup.}
Single-temporal supervision belongs to weak supervision for object change detection.
To investigate the gap between bitemporal supervision and single-temporal supervision, we conducted comprehensive experiments to analyze their performance difference.
The results are presented in Table~\ref{tab:as_bi_st}.
We observe that there is 16$\sim$19\% F$_1$ gap between PCC and bitemporal supervised methods.
Our STAR can significantly bridge the gap to within 10\% when using a large backbone.
And it can be seen that the performance gap keeps getting smaller as the backbone network goes deeper.

\noindent\textbf{Error analysis.}
Comparing Fig.~\ref{fig:error_maps} (e) with Fig.~\ref{fig:error_maps} (d) and (f), we can find that the error of PCC mainly lies in false positives due to various object appearance and object geometric offsets.
This is because PCC only depends on semantic prediction to compare.
To alleviate this problem, that bitemporal supervision directly learns how to compare from pairwise labeled data, while STAR learns how to compare from unpaired labeled data.
From Fig.~\ref{fig:error_maps} (d)/(f), STAR is partly impacted by false positives due to the complete absence of the actual negative samples, e.g. the same object at different times.
Nevertheless, STAR can still learn helpful object change representation to recognize many unseen negative examples successfully.

\begin{figure}
    \begin{center}
        \includegraphics[width=0.48\linewidth]{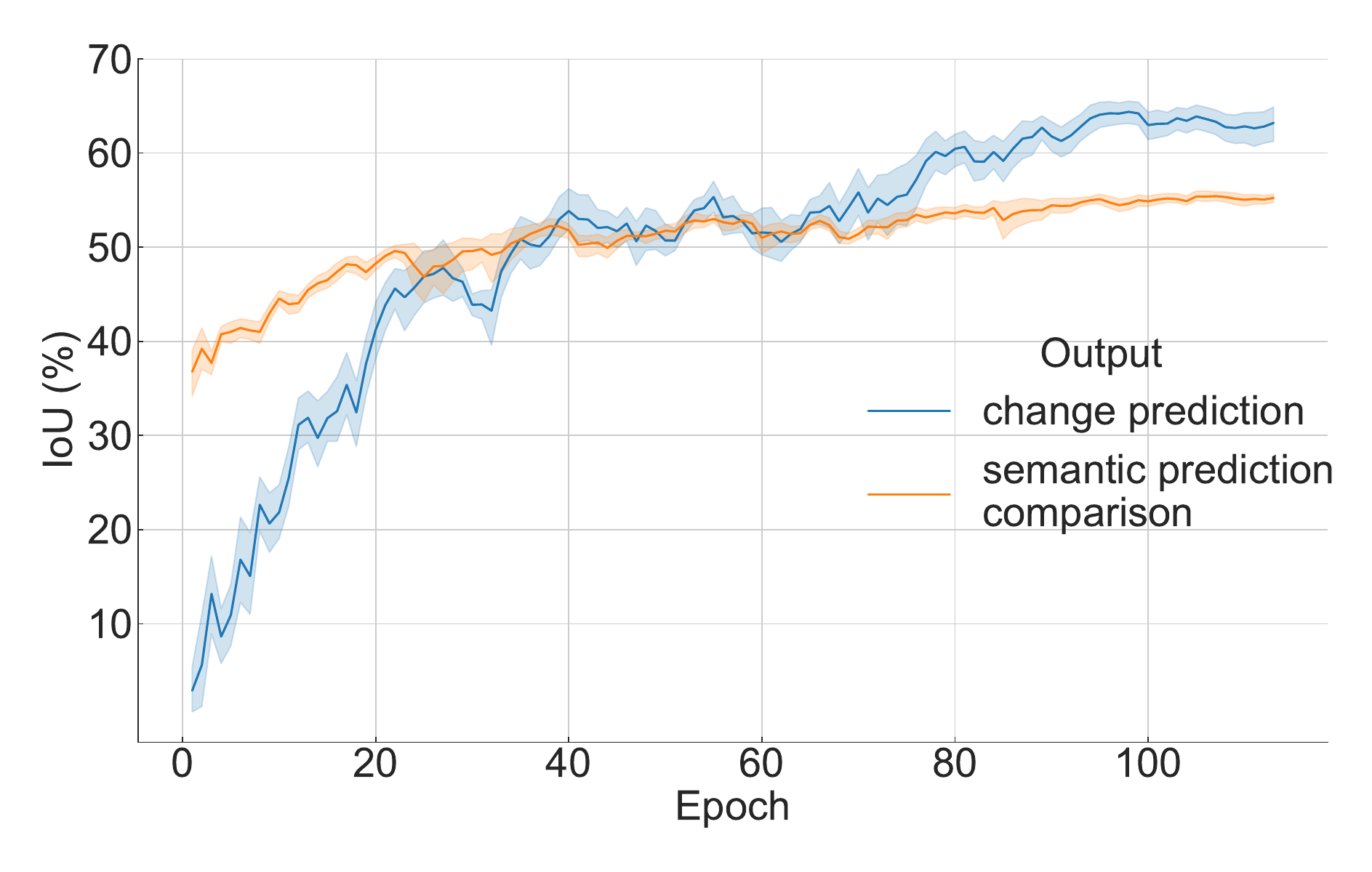}
        \includegraphics[width=0.48\linewidth]{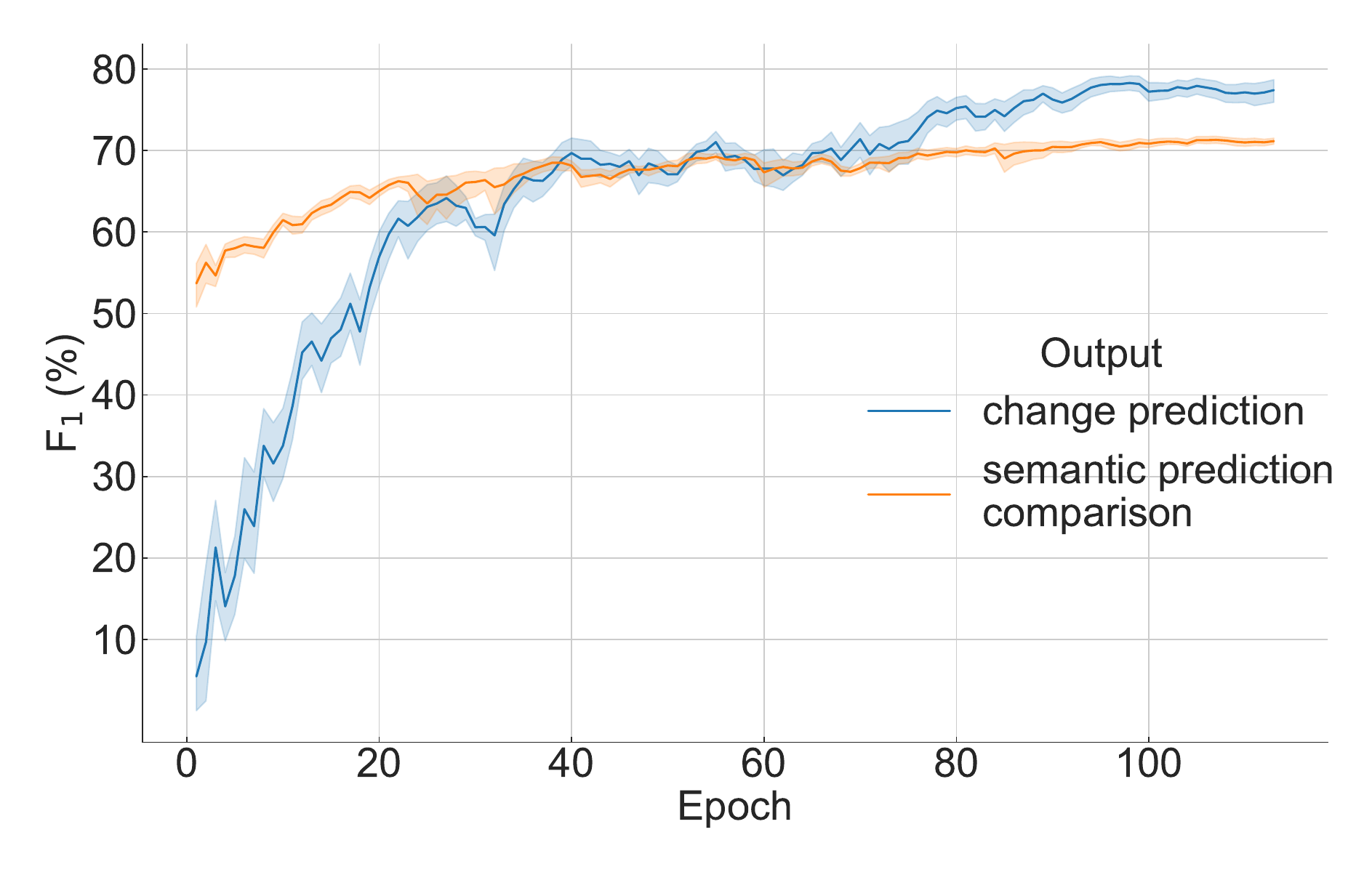}
    \end{center}
    \vspace{-0.2in}
    \caption{Learning curves of IoU (\%) and F$_1$ (\%) on LEVIR-CD$^{\texttt{all}}$ using multi-task outputs from ChangeStar with FarSeg.
        The multi-task outputs include change mask from ChangeMixin and semantic masks from FarSeg.
    }
    \label{fig:multitask_output}\vspace{-4mm}
\end{figure}

\noindent\textbf{Does STAR really work?}
ChangeStar can simultaneously output bitemporal semantic predictions and the change prediction.
The change prediction can also be obtained by semantic prediction comparison.
We thus show their learning curves to explore their relationship, as shown in Fig.~\ref{fig:multitask_output}.
We find that the semantic representation learning has a faster convergence speed than the object change representation learning in ChangeStar.
In the early stage of training ($(0, 40]$ epochs), semantic prediction comparison is superior to change prediction.
This suggests that learning semantic representation is easier than learning object change representation.
In the middle stage ($(40, 60]$ epochs), change prediction achieves similar performance with semantic prediction comparison.
After model convergence, change prediction achieves superior performance than semantic prediction comparison with a large margin.
This observation suggests that STAR can bring extra contrastive information to assist object change representation learning rather than only benefit from semantic supervision.

\section{Conclusion}
\label{sec:conc}
In this work, we present single-temporal supervised learning (STAR) to bypass the problem of collecting pairwise labeled images in conventional bitemporal supervised learning.
STAR provides a new perspective of exploiting object changes in arbitrary image pairs as the supervisory signals.
To demonstrate the effectiveness of STAR, we design a simple yet effective multi-task architecture, called ChangeStar, for joint semantic segmentation and object change detection, which can reuse any deep semantic segmentation architecture via the further proposed ChangeMixin module.
The extensive experimental analysis shows its competitive performances in different domains with cheaper labels.
We hope that STAR will serve as a solid baseline and help ease future research in weakly-supervised object change detection.

\noindent \textbf{Acknowledgements:}
This work was supported in part by the National Key Research and Development Program of China under grant no. 2017YFB0504202, in part by the National Natural Science Foundation of China under grant nos. 41771385 and 41801267, and in part by the China Postdoctoral Science Foundation under grant no. 2017M622522.

{\small
\bibliographystyle{ieee_fullname}
\bibliography{changestar_arxiv}

\begin{thebibliography}{10}\itemsep=-1pt

\bibitem{benedek2009change}
Csaba Benedek and Tam{\'a}s Szir{\'a}nyi.
\newblock Change detection in optical aerial images by a multilayer conditional
  mixed markov model.
\newblock {\em IEEE Transactions on Geoscience and Remote Sensing},
  47(10):3416--3430, 2009.

\bibitem{bourdis2011constrained}
Nicolas Bourdis, Denis Marraud, and Hichem Sahbi.
\newblock Constrained optical flow for aerial image change detection.
\newblock In {\em 2011 IEEE International Geoscience and Remote Sensing
  Symposium}, pages 4176--4179. IEEE, 2011.

\bibitem{chen2020spatial}
Hao Chen and Zhenwei Shi.
\newblock A spatial-temporal attention-based method and a new dataset for
  remote sensing image change detection.
\newblock {\em Remote Sensing}, 12(10):1662, 2020.

\bibitem{chen2019change}
Hongruixuan Chen, Chen Wu, Bo Du, Liangpei Zhang, and Le Wang.
\newblock Change detection in multisource vhr images via deep siamese
  convolutional multiple-layers recurrent neural network.
\newblock {\em IEEE Transactions on Geoscience and Remote Sensing},
  58(4):2848--2864, 2019.

\bibitem{chen2017rethinking}
Liang-Chieh Chen, George Papandreou, Florian Schroff, and Hartwig Adam.
\newblock Rethinking atrous convolution for semantic image segmentation.
\newblock {\em arXiv preprint arXiv:1706.05587}, 2017.

\bibitem{chen2018encoder}
Liang-Chieh Chen, Yukun Zhu, George Papandreou, Florian Schroff, and Hartwig
  Adam.
\newblock Encoder-decoder with atrous separable convolution for semantic image
  segmentation.
\newblock In {\em Proceedings of the European conference on computer vision
  (ECCV)}, pages 801--818, 2018.

\bibitem{daudt2018fully}
Rodrigo~Caye Daudt, Bertr Le~Saux, and Alexandre Boulch.
\newblock Fully convolutional siamese networks for change detection.
\newblock In {\em 2018 25th IEEE International Conference on Image Processing
  (ICIP)}, pages 4063--4067. IEEE, 2018.

\bibitem{daudt2018urban}
Rodrigo~Caye Daudt, Bertr Le~Saux, Alexandre Boulch, and Yann Gousseau.
\newblock Urban change detection for multispectral earth observation using
  convolutional neural networks.
\newblock In {\em IGARSS 2018-2018 IEEE International Geoscience and Remote
  Sensing Symposium}, pages 2115--2118. IEEE, 2018.

\bibitem{daudt2019multitask}
Rodrigo~Caye Daudt, Bertrand Le~Saux, Alexandre Boulch, and Yann Gousseau.
\newblock Multitask learning for large-scale semantic change detection.
\newblock {\em Computer Vision and Image Understanding}, 187:102783, 2019.

\bibitem{fujita2017damage}
Aito Fujita, Ken Sakurada, Tomoyuki Imaizumi, Riho Ito, Shuhei Hikosaka, and
  Ryosuke Nakamura.
\newblock Damage detection from aerial images via convolutional neural
  networks.
\newblock In {\em 2017 Fifteenth IAPR International Conference on Machine
  Vision Applications (MVA)}, pages 5--8. IEEE, 2017.

\bibitem{gupta2019creating}
Ritwik Gupta, Richard Hosfelt, Sandra Sajeev, Nirav Patel, Bryce Goodman, Jigar
  Doshi, Eric Heim, Howie Choset, and Matthew Gaston.
\newblock xbd: A dataset for assessing building damage from satellite imagery.
\newblock {\em arXiv preprint arXiv:1911.09296}, 2019.

\bibitem{he2017mask}
Kaiming He, Georgia Gkioxari, Piotr Doll{\'a}r, and Ross Girshick.
\newblock Mask r-cnn.
\newblock In {\em Proceedings of the IEEE international conference on computer
  vision}, pages 2961--2969, 2017.

\bibitem{he2016deep}
Kaiming He, Xiangyu Zhang, Shaoqing Ren, and Jian Sun.
\newblock Deep residual learning for image recognition.
\newblock In {\em Proceedings of the IEEE conference on computer vision and
  pattern recognition}, pages 770--778, 2016.

\bibitem{hussain2013change}
Masroor Hussain, Dongmei Chen, Angela Cheng, Hui Wei, and David Stanley.
\newblock Change detection from remotely sensed images: From pixel-based to
  object-based approaches.
\newblock {\em ISPRS Journal of photogrammetry and remote sensing}, 80:91--106,
  2013.

\bibitem{ji2018fully}
Shunping Ji, Shiqing Wei, and Meng Lu.
\newblock Fully convolutional networks for multisource building extraction from
  an open aerial and satellite imagery data set.
\newblock {\em IEEE Transactions on Geoscience and Remote Sensing},
  57(1):574--586, 2018.

\bibitem{kirillov2019panoptic}
Alexander Kirillov, Ross Girshick, Kaiming He, and Piotr Doll{\'a}r.
\newblock Panoptic feature pyramid networks.
\newblock In {\em Proceedings of the IEEE Conference on Computer Vision and
  Pattern Recognition}, pages 6399--6408, 2019.

\bibitem{NIPS2012_c399862d}
Alex Krizhevsky, Ilya Sutskever, and Geoffrey~E Hinton.
\newblock Imagenet classification with deep convolutional neural networks.
\newblock In {\em Advances in Neural Information Processing Systems},
  volume~25, pages 1097--1105, 2012.

\bibitem{lebedev2018change}
MA Lebedev, Yu~V Vizilter, OV Vygolov, VA Knyaz, and A~Yu Rubis.
\newblock Change detection in remote sensing images using conditional
  adversarial networks.
\newblock {\em International Archives of the Photogrammetry, Remote Sensing \&
  Spatial Information Sciences}, 42(2), 2018.

\bibitem{mahdavi2019polsar}
Sahel Mahdavi, Bahram Salehi, Weimin Huang, Meisam Amani, and Brian Brisco.
\newblock A polsar change detection index based on neighborhood information for
  flood mapping.
\newblock {\em Remote Sensing}, 11(16):1854, 2019.

\bibitem{mou2018learning}
Lichao Mou, Lorenzo Bruzzone, and Xiao~Xiang Zhu.
\newblock Learning spectral-spatial-temporal features via a recurrent
  convolutional neural network for change detection in multispectral imagery.
\newblock {\em IEEE Transactions on Geoscience and Remote Sensing},
  57(2):924--935, 2018.

\bibitem{peng2019end}
Daifeng Peng, Yongjun Zhang, and Haiyan Guan.
\newblock End-to-end change detection for high resolution satellite images
  using improved unet++.
\newblock {\em Remote Sensing}, 11(11):1382, 2019.

\bibitem{ren2015faster}
Shaoqing Ren, Kaiming He, Ross Girshick, and Jian Sun.
\newblock Faster r-cnn: Towards real-time object detection with region proposal
  networks.
\newblock In {\em Advances in neural information processing systems}, pages
  91--99, 2015.

\bibitem{singh1989review}
Ashbindu Singh.
\newblock Review article digital change detection techniques using
  remotely-sensed data.
\newblock {\em International journal of remote sensing}, 10(6):989--1003, 1989.

\bibitem{tian2020hiucd}
Shiqi Tian, Yanfei Zhong, Ailong Ma, and Zhuo Zheng.
\newblock Hi-ucd: A large-scale dataset for urban semantic change detection in
  remote sensing imagery.
\newblock {\em arXiv preprint arXiv:2011.03247}, 2020.

\bibitem{tran2018closer}
Du Tran, Heng Wang, Lorenzo Torresani, Jamie Ray, Yann LeCun, and Manohar
  Paluri.
\newblock A closer look at spatiotemporal convolutions for action recognition.
\newblock In {\em Proceedings of the IEEE conference on Computer Vision and
  Pattern Recognition}, pages 6450--6459, 2018.

\bibitem{van2018spacenet}
Adam Van~Etten, Dave Lindenbaum, and Todd~M Bacastow.
\newblock Spacenet: A remote sensing dataset and challenge series.
\newblock {\em arXiv preprint arXiv:1807.01232}, 2018.

\bibitem{zhang2020deeply}
Chenxiao Zhang, Peng Yue, Deodato Tapete, Liangcun Jiang, Boyi Shangguan, Li
  Huang, and Guangchao Liu.
\newblock A deeply supervised image fusion network for change detection in high
  resolution bi-temporal remote sensing images.
\newblock {\em ISPRS Journal of Photogrammetry and Remote Sensing},
  166:183--200, 2020.

\bibitem{zhang2017separate}
Xueliang Zhang, Pengfeng Xiao, Xuezhi Feng, and Min Yuan.
\newblock Separate segmentation of multi-temporal high-resolution remote
  sensing images for object-based change detection in urban area.
\newblock {\em Remote Sensing of Environment}, 201:243--255, 2017.

\bibitem{zhao2017pyramid}
Hengshuang Zhao, Jianping Shi, Xiaojuan Qi, Xiaogang Wang, and Jiaya Jia.
\newblock Pyramid scene parsing network.
\newblock In {\em Proceedings of the IEEE conference on computer vision and
  pattern recognition}, pages 2881--2890, 2017.

\bibitem{zheng2020foreground}
Zhuo Zheng, Yanfei Zhong, Junjue Wang, and Ailong Ma.
\newblock Foreground-aware relation network for geospatial object segmentation
  in high spatial resolution remote sensing imagery.
\newblock In {\em Proceedings of the IEEE/CVF Conference on Computer Vision and
  Pattern Recognition}, pages 4096--4105, 2020.

\end{thebibliography}
}

\end{document}